# Explainable AI for Engineering Design: A Unified Approach of Systems Engineering and Component-Based Deep Learning Demonstrated by Energy-Efficient Building Design[1]

Philipp Geyer[2], Manav Mahan Singh[3], Xia Chen[2]

*Abstract*—**Data-driven models created by machine learning (ML) have gained importance in all fields of design and engineering. They have high potential to assist decision-makers in creating novel artifacts with better performance and sustainability. However, limited generalization and the black-box nature of these models lead to limited explainability and reusability. To overcome this situation, we developed a component-based approach to create partial component models by ML. This component-based approach aligns deep learning with systems engineering (SE). The key contribution of the component-based method is that activations at interfaces between the components are interpretable engineering quantities. In this way, the hierarchical component system forms a deep neural network (DNN) that a priori integrates interpretable information for explainability of predictions. The large range of possible configurations in composing components allows the examination of novel unseen design cases outside training data. The matching of parameter ranges of components using similar probability distributions produces reusable, well-generalizing, and trustworthy models. The approach adapts the model structure to SE methods and domain knowledge. We examine the performance of the approach in the field of energy-efficient building design: First, we observed better generalization of the component-based method by analyzing prediction accuracy outside the training data. Especially for representative designs that are different in structure, we observed a much higher accuracy ($R^2 = 0.94$) compared to conventional monolithic methods ($R^2 = 0.71$). Second, we illustrate explainability by demonstrating how sensitivity information from SE and an interpretable model based on rules from low-depth decision trees serve engineering design. Third, we evaluate explainability using qualitative and quantitative methods that demonstrate the matching of preliminary knowledge and data-driven derived strategies and show correctness of activations at component interfaces compared to white-box simulation results (envelope components: $R^2 = 0.92..0.99$; zones: $R^2 = 0.78..0.93$).**

## 1 Introduction

### *1.1 Data-driven prediction in design and engineering*

Data-driven models assist in complex engineering design tasks and are able to sufficiently capture these complexities. Moreover, the computational power required to create models is no longer a limiting factor. Therefore, numerous applications exist in engineering-related domains. Predicting the energy demand of buildings [1]–[5] is important because data-driven models avoid the complex modeling and high computational load required for thermal simulations. These data-driven models are used as surrogates for simulations in designing sustainable buildings [6], [7]. Following the same pattern, surrogate models for structural design and engineering have been created [8]. Extreme learning with reduced training effort has been established for dynamic systems [9]–[11]. Data-driven modeling has been established to predict flows in fluid dynamics [12]. Furthermore, deep learning methods have been applied in operations research and systems engineering for control and decision problems, including the analysis of dynamic systems [13].

An energy-efficient building domain with a dynamic complex thermal system serves as an exemplary domain, representing schemes analogous to other domains of design and engineering. Designers and engineers developing high-performance solutions for such systems require real-time feedback in a process called design space exploration (DSE) to understand how to improve a given design configuration [14]. This exploration process includes variation in design configuration as a key technique for answering what-if questions. Because this process requires the analysis of many variants to obtain this information, the use of physical simulations causes significant modeling and computation loads. This load is substantial for real-time applications when a physical simulation is used, which limits the exploration process. This opens up opportunities for design optimization, as demonstrated in the context of green building design [15], [16]. It not only requires the evaluation of many variants and states but also building energy classification, clustering, and retrofit strategy development [17], [18] as well as control and management problems [19]–[21]. In this situation, data-driven modeling trained on either simulation results or real data

[1] This work has been submitted to Advanced Engineering Informatics on 9 Sept 2024 for potential publication. Copyright may be transferred without notice.

[2] Sustainable Building Systems Group, Institute of Design and Construction, Leibniz University Hannover, geyer@iek.uni-hannover.de

[3] Georg Nemetschek Institute Artificial Intelligence for the Built World, Technical University of Munich

collected from existing artifacts is an interesting alternative to physical simulations.

### *1.2 Limitation of conventional machine learning (ML) and contribution of component-based machine learning (CBML)*

However, there are two major limitations of the application of current data-driven approaches in engineering design.

(1) **Generalization:** Data-driven models are reliable only within the distribution of the training dataset and often show limited generalization. Techniques such as regularization and hyperparameter tuning have been developed to improve model generalization [22]. However, owing to the nature of design and engineering, creating novel artifacts requires robust models with reliable and guaranteed generalization. Decision-makers need to be sure that the models will predict correctly in unseen cases that differ from the training data. Most application cases mentioned above vary the parameters without changing the model structure, an approach that we call the monolithic model. This parametric monolithic approach allows the treatment of generalization as a statistic of the input parameters that identify the model boundaries. However, a pure parametric model is also significantly limited in generalization, because novel design cases frequently have different structures that cannot be covered with parametric changes; consequently, the adaptation of the features of the data-driven model remains incomplete, leading to inaccurate predictions. A slight mitigation uses characteristic numbers instead of the direct use of design parameters [23], —a best practice that serves as a benchmark in Section 3.

(2) **Explainability:** Furthermore, designers and engineers need insights into how the models predict plausibility, approve and justify results, and gain an understanding of the behavior of the design configuration and the nature of the design space. This requires the explainability of artificial intelligence (XAI), particularly for data-driven models applied in engineering design. To address the shortcomings of the black-box character of such models, considerable research has been conducted to develop XAI [24]–[26]. On the one hand, there is a limited set of *interpretable models*, also known as white-box models, whose elements humans can read directly, in contrast to the explainability of predictions, because readable terms need to be constructed using additional methods [27]. This is also called model-based interpretability, in contrast to post hoc methods [28]. Direct readability relies on connecting model elements with semantics or meaning and allows for a direct understanding. Linear regression is an old technique used in data-driven models with direct interpretability [29]. Such methods have high potential in engineering because their terms are meaningful to engineers when applied to improve designs, as shown in the energy domain [30], [31]. However, a huge set of *post-hoc methods* adds explainability to the predictions of ML black-box models. Such XAI methods include visualizations, model simplifications, text explanations, local explanations, feature relevance, and explanations by example [24]. Local interpretable model-agnostic explanations (LIME), which constructs local explanations and SHapley Additive exPlanations (SHAP), which shows feature relevance based on game theory, are two important post hoc methods that have also been transferred to data-driven energy models in several cases [32], [33]. Furthermore, conventional engineering methods such as sensitivity analysis (SA) help explain the predictions of data-driven models [34]. All these methods focus on—and this is a current gap between data-driven models and engineering—the relevance of input features in predicting the output. However, the intermediate steps required to derive the output were not revealed. In a monolithic network, understanding the intermediate steps involves examining the activation of the hidden layers. Our experiments on energy efficiency and those of others have shown that directly accessing activations in conventional deep neural networks (DNNs), as well as using SHAP, deliver only limited explainability [35]–[37]. For instance, in energy-efficient building design, even if relevant features are discovered, detailed information about heat flows over time allows engineers to understand what causes the heating or cooling demand to be unavailable by the methods described above. Therefore, direct model interpretability, that is, the readability of model content by humans, is of vital interest.

**Contribution of CBML:** In contrast to monolithic models, we developed *component-based machine learning (CBML)*, a method that comprises a model architecture for individual prediction following a systems engineering approach. In contrast to the existing systems engineering integrations of ML [38]–[41], CBML does not use a monolithic model but breaks the data-driven model down to the design components. Ontologies provide a background for such system decomposition, such as existing schemes [42]–[45], which have been combined with ML, especially in the energy context [46], [47]; however, data-driven models have not yet been organized according to ontologies.

As a consequence of this system decomposition, modeling aligned with the design case allows for a far broader range of configurations, providing much better *generalization*, that is, higher accuracy in design cases that are less similar to the training data, as demonstrated in Section 3. Using pretrained component models in new contexts and thus modifying the model architecture, CBML has some parallels with transfer learning (TL) [48], [49]; in the case of artificial neural networks, the model forms a DNN. However, in contrast to layers being reused in TL, the interfaces between components have engineering quantities with units, and therefore provide interpretable information. This information enables natural interpretability and understanding by designers and engineers and therefore supports the *explainability* of predictions. In the case of a DNN, the network activations are interpretable as engineering quantities at the component interfaces. In Section 4, we demonstrate these benefits in terms of explainability and evaluation techniques [50], [51] to further understand the design space.

The first publication of CBML in 2018 provided a proof of concept [52]. In the following years, we used this method in the context of research on multi-level-of-detail modeling [53] to predict the yearly totals of energy consumption and to examine information requirements under uncertainty [36], [54]. We examined time series predictions and their robustness against sparse data [55]. The remainder of this paper reviews the generalization capabilities and focuses on the explainability potential of this method. Section 2 explains the CBML approach in detail in the context of systems engineering. We review the different aspects of its generalization capabilities and summarize the results of our work in Section 3. In Section 4, as the main contribution of this study, we address the gap in XAI for engineering, which consists of nontransparent unitless numbers caused by the black-box character of many models derived by machine learning. We explore the potential of engineering quantities with units at the component interfaces. This includes not only the introduction of different concepts and methods to understand individual prediction, but also the general behavior of the design with its inherent design space to be explored to support the engineering decision process. Finally, we evaluate explainability using standard methods from the XAI domain.

## 2 CBML methodology and its evaluation

To improve both the generalizability and explainability, we developed a component-based approach that aligns data-driven modeling with design and engineering. We created data-driven models for design and engineering processes following a systems engineering paradigm described in [56]. Decomposition according to this paradigm using system components with input and output parameters forming the interfaces between the components is the key element of CBML, and the above-mentioned ontologies serve as the background for decomposition. This approach provides a basis for engineering the interpretability and reusability of components, leading to broad generalization. Furthermore, it makes data-driven models accessible to existing system analysis methods such as those compiled by Kreimeyer and Lindemann [57].

The key features of CBML are the setup and training of data-driven models at the component level and prediction at the system level, as shown in Figure 1. The result of the component training is a component set for a specific use case, for example, the prediction of the energy demand for a specific type of building. Prediction occurs by modeling the design as a system of components and running the prediction at the system level.

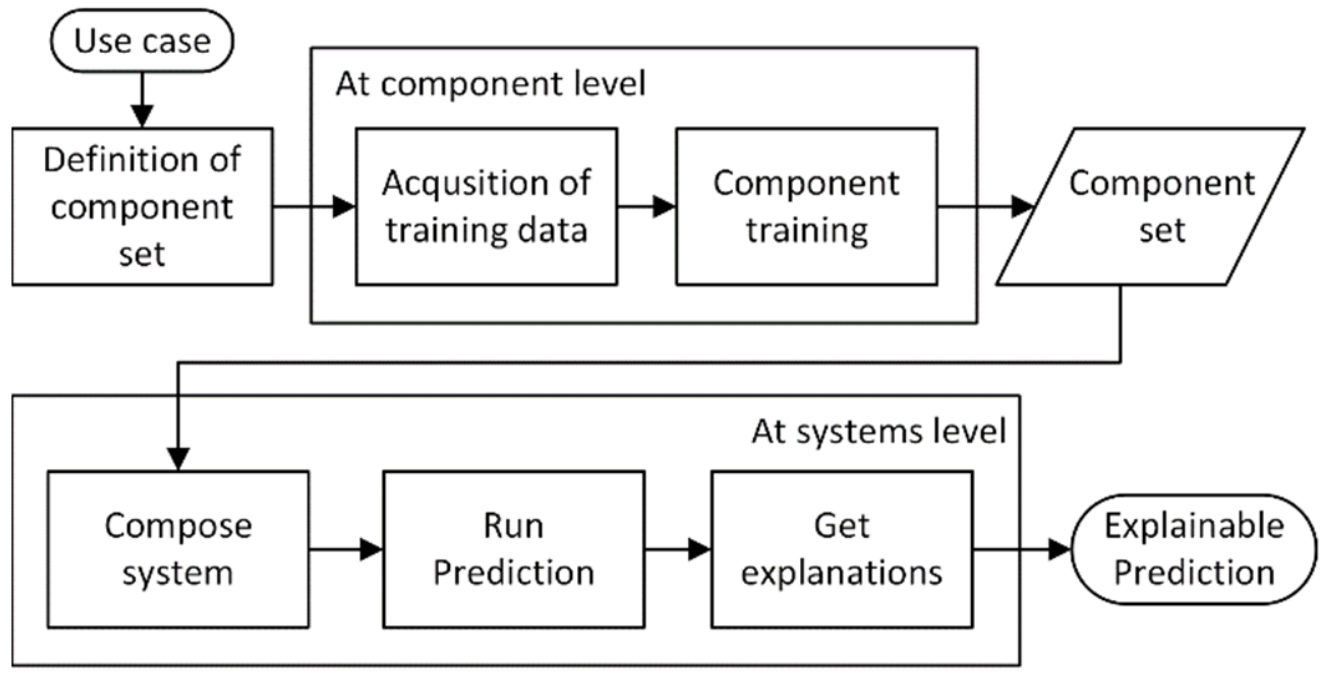

**Fig. 1. Flow chart of the CBML process.** Component training occurs at the component level and prediction at the system level.

To construct the component models, we used artificial neural networks (ANNs) because of their high flexibility and capacity, although other ML modeling methods are also possible. The data-driven model based on ANN methods forms a DNN using the components' composition with one particularity: In addition to the layer functions $f_L$, the component function $f_c$ includes the interfaces' functions for scaling $\nu$ and inverse scaling $\nu^{-1}$ to provide the interpretable engineering quantities $\mathbf{y}_r$ with units:

$$f_c(\mathbf{x}) = \mathbf{y}_r = \nu^{-1}\left(f_l^n\left(\dots f_l^2\left(f_l^1\left(\nu(\mathbf{x})\right)\right)\right)\right) \text{ with} \tag{1a}$$

$$f_l(x) = \mathbf{y}_l = \sum_{i=1}^{n} \mathbf{w}^T \phi(\mathbf{x}) + w_0; \tag{1b}$$

$$\nu = \frac{x - x_{\min}}{x_{\max} - x_{\min}}; \tag{1c}$$

$$\nu^{-1}(y_s) = y_s\left(y_{\max} - y_{\min}\right) + y_{\min}. \tag{1d}$$

where $\varphi$ is the activation function, $\mathbf{x}$ are the input parameters, and $\boldsymbol{w}$ and $w_0$ are learning parameters. The linear scaling function $\nu$ converts and normalizes the engineering quantities to a unitless number between 0 and 1. Inverse scaling reverses this normalization at the outgoing interface of the component to reconstruct an engineering quantity with units. Both scaling and inverse scaling were constructed during component training depending on the parameter ranges.

As a consequence of the scaling and inverse scaling, during composition of the data-driven model for prediction, shown in Equation 1a, all inputs $\mathbf{x}$ and outputs $\mathbf{y}$ of a component are subject to engineering units. Therefore, the activations at the interfaces between the components have units of heat flow, heating and cooling demand, and final energy consumption in W, kWh/a, or kWh/m$^2$a in the case of energy performance prediction. Thus, a DNN is created that allows the direct interpretation of engineering quantities within the network.

During the training process, labeled training data for a supervised learning procedure were used at the component level. These data form the training set $X_c$ for each component $c$ with $n_c$ samples to connect the features $\mathbf{x}_c$ with labels $\mathbf{y}_c$ to act as the ground truth.

$$X_c = \left\{ \mathbf{x}_c, \mathbf{y}_c \right\}_c^{n_c} \tag{2}$$

Training data were collected from real-world measurements [58] and synthetic data generated by state-of-the-art dynamic simulation tools [52], [59]. Training takes place in a component-by-component supervised learning process, using the respective input and output data, as illustrated for the window and wall components in Figure 2a. Given such labeled sets, standard training processes serve to adapt the component functions, for example, by adapting the weight parameters $\mathbf{w}$ and $w_0$ in Equation 1 using a gradient-based method.

For prediction, the components are composed to represent the design artifact in its current configuration. By connecting the inputs and outputs of the components to a system representing new designs, data-driven models can be reused for unseen configurations (Figure 2b). To illustrate the composition for prediction, Figure 3 shows two examples of partial component structures. The first structure is that used in the training data; the second deviates from the training data and allows the examination of generalization under structural variation. Aligning the component structure with digital modeling, building information modeling (BIM) [60] and industry foundation classes (IFCs) [29] facilitated the automatic generation of data-driven models. Figure 2 provides a class diagram of CBML for energy demand prediction and Figure 3 shows partial instance diagrams.

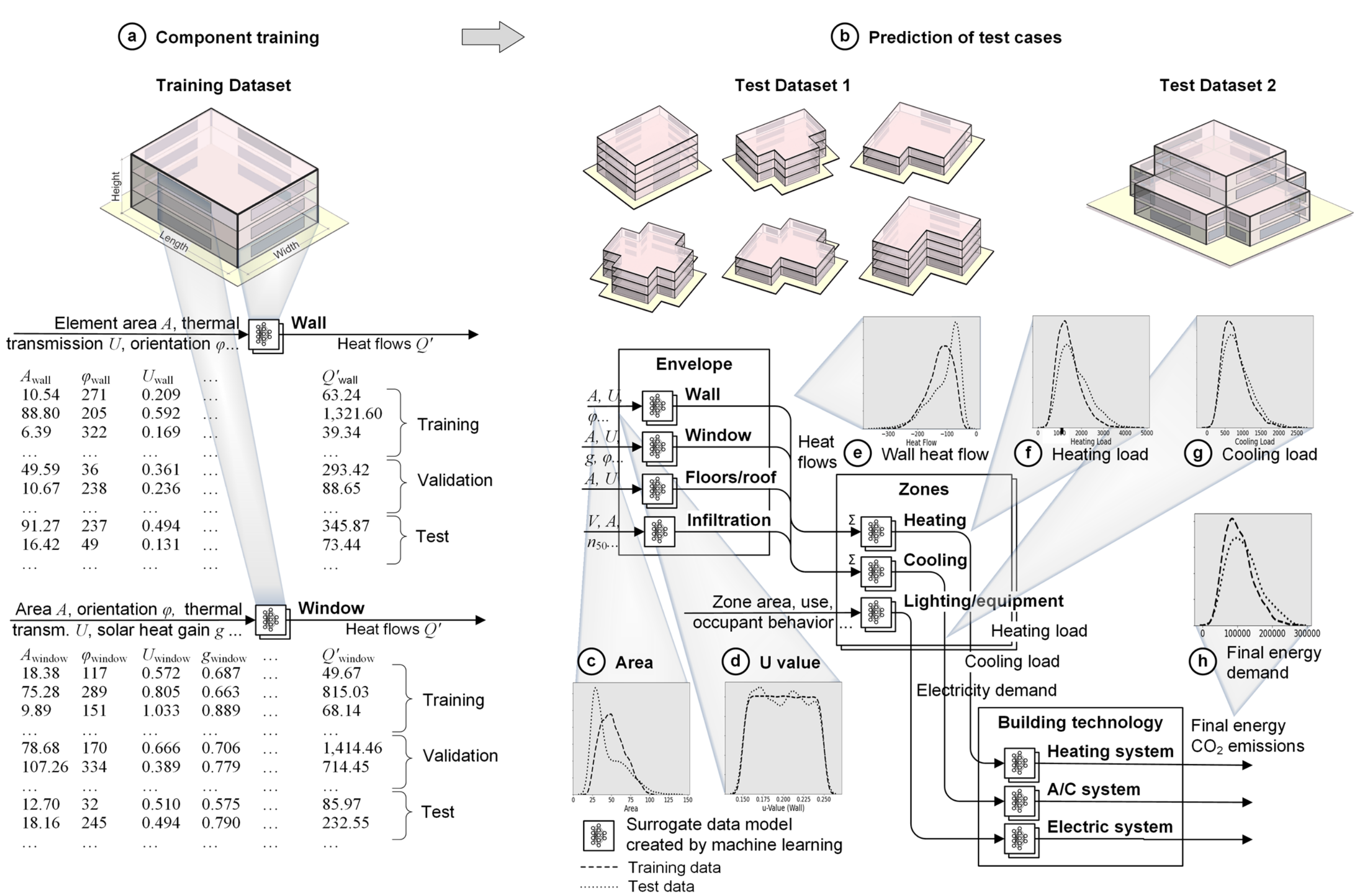

**Fig. 2. Component-based machine learning (CBML).** (**a**) Training occurs at the component level. (**b**) To predict new cases, the components are composed of new system configurations according to the design structure. This allows for the representation of novel unseen configurations beyond the training data based on domain knowledge. Simultaneously, it enables the matching range and probability distribution of the training data shown for exemplary parameters in the gray boxes (light dotted line: training data; thick dashed line: test data).

Formally, in prediction, the system function $f_s$ as a DNN is composed of $h$ hierarchical levels with nested component functions $f_c$ with $n_h$ components at each level:

$$f_s\left(\mathbf{x}_s\right)=\hat{\mathbf{y}}_s=f_c^{h,\,n_h}\begin{pmatrix} & f_c^{2,\,1}\begin{pmatrix} f_c^{1,\,1}\left(\mathbf{x}_{1,1}\right) \\ \cdots \\ f_c^{1,\,m}\left(\mathbf{x}_{1,m}\right) \end{pmatrix} \\ \cdots & \cdots \\ & f_c^{2,\,n_2}\begin{pmatrix} f_c^{1,\,m+1}\left(\mathbf{x}_{1,m+1}\right) \\ \cdots \\ f_c^{1,\,n_1}\left(\mathbf{x}_{1,n_1}\right) \end{pmatrix} \end{pmatrix} \tag{3}$$

The validation occurs at the system level. For this purpose, a labeled validation dataset $X_s$ with various $s$ structures, that is, differing component compositions, was used to examine the performance of typical system structures. This set includes the system topology as a directed acyclic graph $G_s$, which includes the components used as nodes $V_c$ and their connecting edges $E_c$ Second, it includes the parametric features $\mathbf{x}_s$ at the system level with labels $\mathbf{y}_s$ to act as the ground truth for validation.

$$X_s=\left\{G_s,\mathbf{x}_s,\mathbf{y}_s\right\}_s^{n_s}\ \text{ with } G_s=\left\{V_c,E_c\right\} \tag{4}$$

On the basis of $G_s$ and $\mathbf{x}_s$, the matrix in Equation 3 is populated and delivers predictions $\hat{\mathbf{y}}_s$ to be compared with the ground truth $\mathbf{y}_s$ in the validation dataset $X_s$. This validation provides information on the overall system prediction performance of the data-driven model.

This approach, which is based on systems engineering and components, provides a much wider generalization than the monolithic approach. The structure of data is met by selecting and composing data-driven models as required to represent the design configuration. In the example, the monolithic model based on the training data shown in Figure 2a would only allow prediction for box buildings because of the structure of the data and the model, and the component-based approach allows prediction for a broad variety of design cases, as exemplified by the configurations shown in Figure 4, top. Because the components occur in similar configurations in the novel design cases and are present in the training dataset, the method avoids extrapolation, which is a serious problem for data-driven models. The gray boxes in Figures 2c-h show examples of the matching of parameter ranges and distributions as histograms. There are parameter ranges, such as the u-value, that determine the heat transmission through walls and windows, which can be configured directly and thus show a perfect match (Figure 2d). Many other parameters depend on previously determined parameters and predictions; therefore, no direct control is possible, but the match depends on the other parameters of the design configuration. For instance, the wall area (Figure 2c) depends on the building shape and heating and cooling loads (Figures 2f, g), and the final energy demand (Figure 2h) depends on the dynamic thermal processes determined by the component heat flows of the building envelope and zone parameters. Therefore, only an approximate match is achievable. In general, all the data showed a good match for the example, although the configurations of the training and prediction were significantly different.

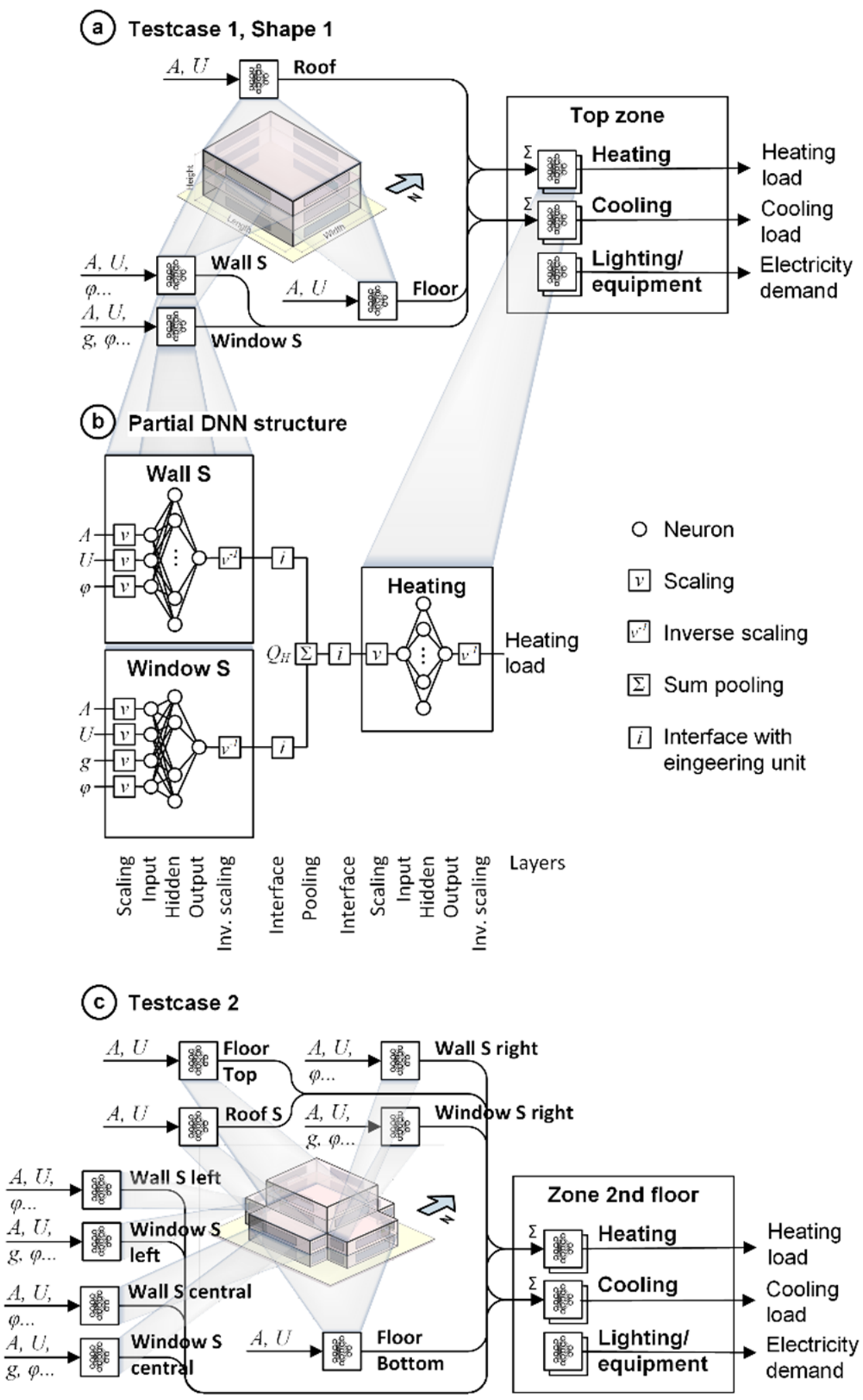

**Fig. 3. Instance diagram of Test cases: (a)** The instance of the top zone of Test case 1, Shape 1 exemplifies the composition of a box-like design for prediction; **(b)** the partial structure illustrates the generation of a DNN for this shape; **(c)** the composition for Test case 2 shows how components are used, which is a structure different from training for prediction. For readability, only one zone and only the south façade, roof, and floor of this zone are shown in all subfigures.

This demonstrates that CBML complies with two preconditions for the use of data-driven models: (1) the structure of the model matches, that is, the parameters and their meanings coincide in training and prediction, and (2) the parameter distribution in training covers the distributions in prediction. By allowing this matching with a varying model structure, CBML delivers superior generalization with high flexibility. Appendices A and B describe the methodological details of the experimental data generation and the monolithic baseline model used as the benchmark. Appendices C and D describe the creation of the component-based models for single values and time series, respectively.

To summarize, the scheme to achieve this generalization consists of the extraction of training data from one case, encapsulation of component behavior as a data-driven model by ML, and application of the composition of data-driven models to novel structures. This scheme is a form of inductive TL [48] that aligns engineering reasoning to the underlying rules and, eventually, to basic physical laws as a form of domain knowledge. The matching of the structure of the domain knowledge and that of the probability distributions of features or parameters forms important conditions for the transfer [49]. The component approach provides the basis for inductive TL. We see this as a special class of inductive TL that extends data information using engineering knowledge consisting of the decomposition and recomposition of artifacts according to the paradigm of systems engineering.

Training and test datasets with different characteristics were generated to evaluate the generalization capabilities. The training process was based on the simulation results from the box building shown in Figure 2a; the details are described in Appendix A. Two test datasets with different design structures allow a realistic evaluation of the generalization capabilities. Randomly generated design footprints form the first test set (Figure 2b, Test Dataset 1). This set represents typical options in the early design phase of a building, and includes a set of shapes that are different from the box building in the training data. As a constraint on random generation, all stories of a building have the same shape, which matches many real-world buildings. The second set (Figure 2b, Test Dataset 2) is based on a designed building configuration that is more complex because not all stories have the same geometry; roofs occur at different levels, and the zone at level two is connected to the roofs and a floor slab at the top. All generated models in the training and test datasets were parametric in terms of geometry and engineering properties, which reflected the design space at an early level of preliminary design. Section 3 provides the results of the generalization capabilities.

Therefore, both test datasets contain additional complexity compared with the training data and serve to test the ML models for their ability to generalize beyond the training data. For the evaluation, the predictions of component-based ML models and a monolithic model based on state-of-the-art baseline methods were compared. Because transfer between training and test datasets is not possible based on their models' different structures, the state-of-the-art does not use geometric parameters or zone information directly, but rather performance-characterizing numbers valid for all buildings, such as floor area, height, number of floors, and relative compactness, following the methods of Chou and Bui [23]. The use of such characteristic numbers currently delivers the best possible results for monolithic models in representing arbitrary design configurations (for details on the baseline method, see Appendix B).

To examine and demonstrate explainability, a common approach is to analyze activations and their propagation in a selected prediction case [61]. Layer-wise relevance propagation (LRP) [62] is a method used in image classification. Similar to the DNN used in image analysis, the CBML structure forms a DNN. To evaluate the capabilities of explainability, we composed prediction models for Test Dataset 2 according to the CBML method, which includes inverse scaling by the function $v^{-1}$ at each interface and applied selected interpretation methods. First, it is straightforward to examine absolute quantities. In Section 4.1, we present the results of the interpretation of these quantities, which enables engineers and designers to understand and verify these results using domain knowledge. An alternative approach to directly examining activation is local variation according to DSE and sensitivity analysis to gain information on the reasons for prediction. We used this approach, which is a traditional engineering method for understanding model behavior and is closely related to LIME [63], [64]. It is a model-agnostic post hoc technique for explainability [24]. From an engineering perspective, a linear local model built on DSE results in a linear regression coefficient that describes sensitivity and enables the interpretation of the importance of the parameters in the model. Sensitivity analysis as described by Menberg et al. [65] serves to generate the mean absolute mean value $\mu^*$ of the elementary effects (EE) with an $\Delta_i$ of 5% (for details see Appendix E). Section 4.2 provides exemplary sensitivities and explanations.

Another method for gaining insight into the reasons for the prediction is the use of local surrogate models. We used decision trees as a local surrogate model, which form a post hoc method of explainability [24]. Trees with low depths allow the extraction of rules that are understandable by engineers and provide further links to domain knowledge. Appendix F describes the methodological details of tree generation, and the results in Section 4.3 show exemplary derived rules.

Finally, we evaluated the explainability by interpreting the predictions of the CBML approach, which offers user insights and thus helps create trust. Schemes for evaluating explainability developed by [24], [50], [51], [66] serve as the background. Using the scheme of Nauta et al. [51], we first use a white-box test to evaluate information at component interfaces, examine explainability capabilities, and ensure correctness; second, we check coherence with users' knowledge by evaluating statistics of interface values against design strategies as relevant domain knowledge. Section 4.4 shows the results of the evaluation.

### *2.1 Domain knowledge and its integration in CBML*

Component-based ML aligns data-driven models with the domain knowledge. In the case used in this study, this knowledge focused on the complex dynamic thermodynamic processes between the building, its environment, and its control for a comfortable indoor environment with low energy demand. To enable readers unfamiliar with this domain of knowledge to follow our methods, we provide a concise introduction in this paragraph. Several key factors are involved in the construction of energy-efficient buildings. First, the area and thermal insulation of the walls, windows, roof, and floor slabs, the so-called u-value in $W/m^2K$, determine the heat flow from indoor to outdoor space. The envelope areas of these elements, which are responsible for heat loss, strongly depend on the shape of the building. Therefore, compactness, that is, the ratio of the façade area to volume, plays an important role and serves as an characteristic number. Second, the solar transmissivity of the windows, the so-called g-value, determines how much heat energy the sun irradiation generates in the building, replacing heat from the building system in winter, but causing a cooling demand in summer. Heat capacity (J/kgK) plays an important role in determining the amount of surplus heat stored in building components, thus describing the dynamic interaction between solar gains, internal gains (heat emitted by users and devices), and heating and cooling systems. Given these dynamically interacting factors, it is complex and not trivial for designers and engineers to understand the thermodynamic behavior of a given design configuration and find a solution that performs well.

The taxonomy proposed by von Rüden et al. [67] lists four entities in a prototypical ML pipeline for integrating knowledge: training data, hypothesis sets, learning algorithms, and final hypotheses. The focus of CBML is the integration of domain knowledge in the hypothesis set by radical organization of the model architecture according to domain knowledge and respective structures. In the case of the example domain, i.e., energy-efficient building design, the structure originates from the system of the building, including its services and constructions, as described in the mentioned domain ontologies. The domain-related model structure enables further connection to domain knowledge in the final hypothesis, i.e., the resulting predictions of CBML, and eventually generates explainability through interpretation, as discussed by Beckh et al. [68]. Another source of knowledge embedded in these components is the use of dynamic simulations to generate training data. By acquiring and using simulation results aligned to the component structure, the CBML incorporates the domain knowledge embedded in the complex system of the dynamic simulation model and its incorporated differential equations.

### *2.2 Domain characterization and method transfer*

This study provides evidence for the benefits of the CBML method using energy-efficient building design as a test domain, focusing on the thermodynamic behavior of buildings and the dynamics of their heating and cooling systems. However, we expect that this method will also be applicable to other domains involving systems with similar characteristics. The following list aims to foster this transfer by explicitly describing the characteristics of the test domain.

- The test domain has partly linear or linearized and partly nonlinear ordinary or partial differential equations (ODEs/PDEs), as well as differential algebraic equations (DAEs) [69]. Simulation environments applied to these equation systems, such as the EnergyPlus software [70] used to generate data, involve solvers.
- The hierarchical nature of the system is a second characteristic of the test domain. Typically, the configuration of a building envelope determines the heating and cooling demands of the zones. Heating and cooling systems perform dynamically, depending on this demand. Organization as a data-driven system with such a unidirectional information flow cuts some of the dynamic feedback loops; however, our results show that this is less relevant than the missing adaptation to the specific design configuration enabled by the components.

Other domains dealing with systems have similar characteristics, which suggests the transferability of the method. Examples include systems in electric circuits and their respective control as well as power engineering problems with typically nonlinear ODEs and DAEs [71], multibody mechanics that involve ODEs and DAEs [72], and structural engineering and dynamics [73]. Depending on the number and importance of loops, we expect CBML to be applicable, although this has up to now not been proven.

## 3 Results for generalization

The quantitative results in terms of generalization tested by the set of random shapes demonstrate the higher precision of the component-based method compared to the state-of-the-art monolithic method. The deviation of the ML prediction from the simulation results, which acted as the ground truth, showed a mean absolute percentage error (MAPE) of only 3.76% and an $R^2$ of 0.99, whereas the monolithic baseline model showed a MAPE of 5.88% and an $R^2$ of 0.94 (Figure 4a). The second set of design configurations adds more complexity, similar to what is usually present in realistic design and engineering cases of buildings. Under these conditions, the advantage of the component-based method over the monolithic method significantly increases. The MAPE of the component-based method is 4.96%, and $R^2$ is 0.94, whereas the MAPE of the conventional method is 12.72%, and $R^2$ is 0.71 (Figure 4c). The monolithic model underestimates energy demand, which is problematic from a domain perspective, whereas the component-based approach does not and is more precise (Figure 4). This increase in the difference for more complex cases indicates that the components show a much better ability to generalize the typical design configurations of a domain instead of being linked to the behavior observable in the training dataset and the related case. Further experiments have shown better generalization of the CBML compared to the monolithic method [55]. We systematically reduced the number of available training samples and observed a decrease in the accuracy by tracking the test error. The results shown in Figure 5 prove a slower drop with increasing sparseness of data for CBML compared to monolithic modeling. This observation provides evidence of robustness in the case of sparse data and demonstrates the ability of CBML to generalize better.

## 4 Results for explainability

Using the derived data-driven model and the representative case based on Test Dataset 2 (Figures 4c, d), we demonstrate different explainability approaches enabled by component-based ML. The first subsection addresses the intrinsic explainability offered directly by interpreting the component interfaces in the system of the data-driven model. The second subsection deals with sensitivity analysis as a form of engineering interpretation of DSE locally around an interesting configuration, which is the representative case shown in Figure 4d. The third subsection presents a decision tree as an interpretable surrogate model based on local DSE data, and derives engineering rules that match the domain knowledge of design and engineering. In the final subsection, conclusions regarding the explainability of these models are evaluated against domain knowledge.

### *4.1 Engineering insights in component systems*

Figure 6 shows exemplary predictions at the interfaces to illustrate how the CBML model offers engineering insights thatallow for explainability. The first examples are the predictions of the yearly averages and totals (6a, e, h) and time series (6b, f) for wall heat flows, zone cooling loads, and energy use intensity (EUI). The yearly numbers allow designers and engineers to identify parts of a building that cause high energy demand. For instance, the time series predictions (6b) show specific dynamic behavior, allowing for an understanding of high demand, such as the heat flowing through the east walls of the representative test case building. In July, the morning sun heats the wall, causing peaks in each of the daily flows, for example, indicating the potential of thermal energy usage, whereas the medium-level flows show conduction of heat from the outside air to the indoor space, causing cooling loads. Negative values indicate heat rejection through the walls at night, which reduces cooling loads. Furthermore, the dynamic behavior of the cooling systems and total demand (6f, i) show profiles caused by the interaction of occupancy and the system, such as internal gains and system shutdowns on weekends. In cooling loads (6f), the consequent high peak after the weekend compared to the Friday before the weekend is predicted correctly; this is usually caused by the heating-up the building while the cooling system is turned off over the weekend.

We emphasize that this information is obtained from interpretable internal activations of a DNN formed by the component system. Although this is purely activation of a DNN prediction, interpretation of this information provides explainability that allows domain experts to understand processes, check results for plausibility, and evaluate and modify the current design to anticipate improved behavior.

Furthermore, rather than using only one configuration, comparing and analyzing multiple configurations, such as given by sensitivities (Figures 6b, f, i) and decision trees (Figure 6d), provides further information for DSE. Subsection 4.2 demonstrates sensitivities in detail and Subsection 4.3 deals with trees and derived rules.

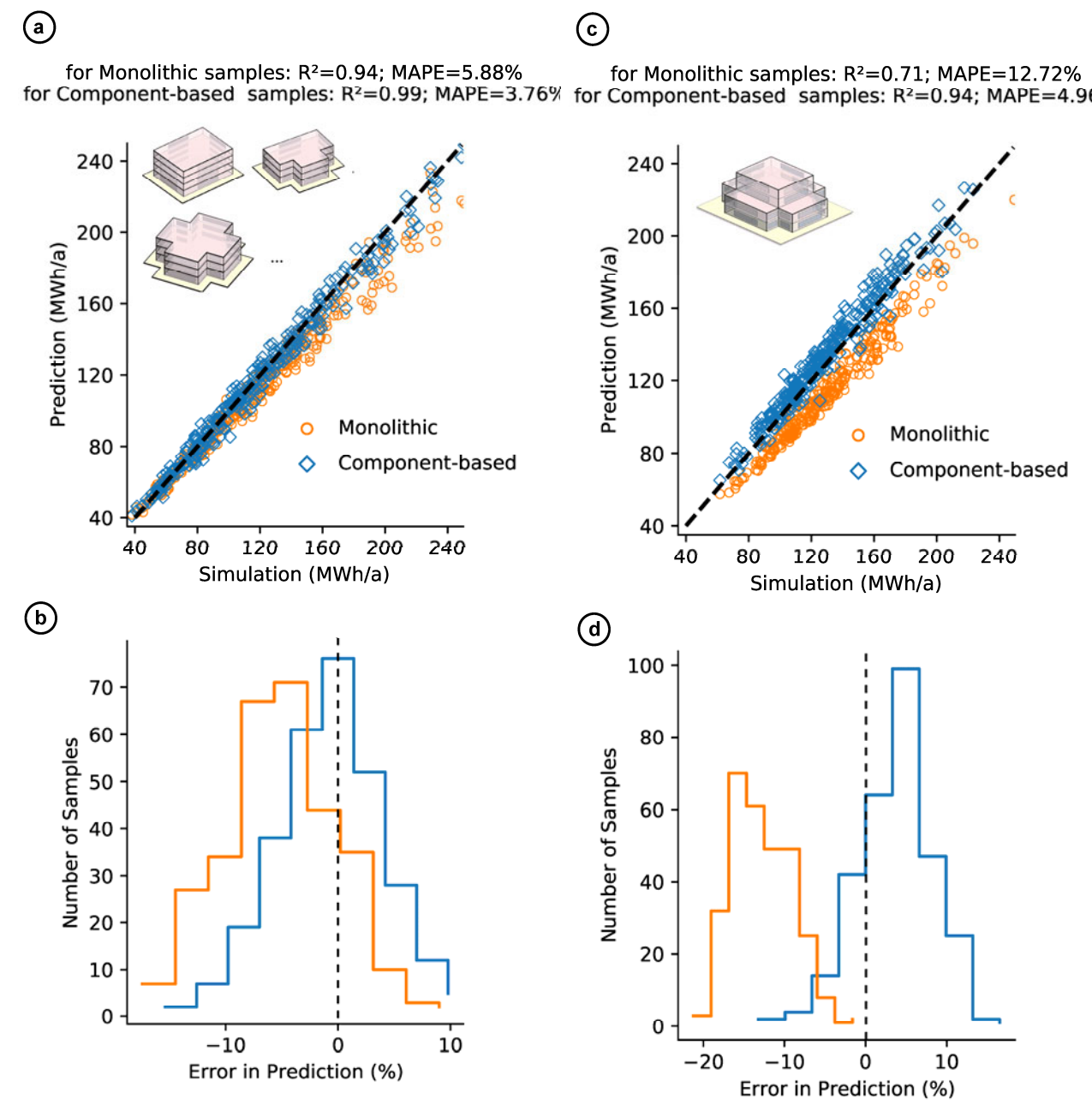

**Fig. 4. Prediction accuracy of test cases based on the ML components to evaluate generalization compared with state-of-the-art monolithic ML models.** The comparison of ML predictions by the component-based method and state-of-the-art monolithic methods to ground truth by simulation serves to examine generalization. **(a), (b):** Comparison of randomly generated shapes with all stories having the same geometry as in the training data; **(c), (d):** more complex design configurations varied by engineering and geometry parameters. This configuration differed from the training data for a roof at the intermediate level.

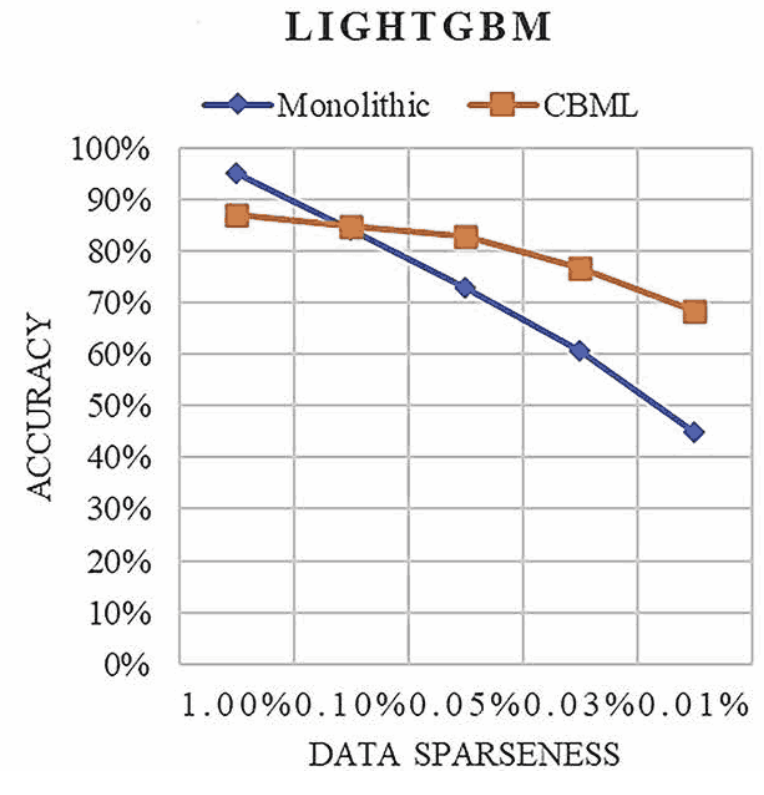

**Fig. 5. Robustness against sparse data.** CBML exhibits significantly higher robustness against data sparseness, providing evidence of its generalization capabilities **[55]**.

### *4.2 Local model explanation by sensitivity analysis*

Figure 7 shows the selected sensitivities for a representative test case. First, the matrix shows high sensitivity of the south wall and window heat flow to the length and height of the building. A change of these parameters from ±5% leads to about 100 $W_{avg}$ additional heat loss but about 300 $W_{avg}$ additional heat gains through the windows (Figure 7a). This tells domain experts that south windows are worth considering for heat gains and energy savings. However, to understand the actual potential, it is necessary to determine whether these heat gains occur in the summer or winter. The g-value, also called the solar heat gain coefficient, provides an answer (Figure 7b): a change in the g-value increases the heat gain by 400 $W_{avg}$, reduces the heating load by 380 $W_{avg}$ and the total operational energy by 250 kWh/a, and increases the cooling load by 370 $W_{avg}$. These observations indicate that heat gains support the heating system in winter and that external shading is a good option in summer. This exemplary interpretation of the behavior of the south windows in the representative case shows how such information provides insights and helps domain experts draw conclusions for design development. In addition to direct engineering reasoning, sensitivity is a means of determining the design parameters that are important. In early design phases, they provide an indicator of which decisions should be made early to reduce the uncertainty in predictions, and thus offer potential in guiding decision-makers through the process [54].

The component-based structure of the model and the calculation of sensitivities make system analysis and complexity metrics available [57]. This connects data-driven models to the design structure matrix (DSM) approach, which deals with the structure of design artifacts, processes, and teams and aims at optimal management of dependencies [74]. As an example of such a technique, Figure 9 shows the sensitivity matrix of the key variables and main internal parameters, extending the information shown in Figure 7. Analyzing the matrix reveals clusters, such as strong geometric dependencies at the top of the matrix, the window g-value and u-value discussed previously, a cluster of operations linking office hours, heat gain of equipment and light and occupancy to loads, and energy demand (last four rows and columns of the matrix). This directs decision-makers to parameter groups that need to be considered simultaneously.

Moreover, summing the columns and rows in the matrix provides activity and passivity, which are two common metrics of system variables [57]. Activity (Figure 9b) indicates the variables that have a high potential to control the system. In this example, the matrix indicates the important role of the building geometry. The passivity (Figure 9c) reveals parameters that have strong reactions and thus strongly depend on the configuration of the system. Among these parameters, the cooling load is striking, which means that the system in its current configuration is relatively sensitive to cooling loads, and there is a high potential to improve the performance by looking at this parameter and its influencers.

In summary, the calculation of sensitivity as a means of understanding dependencies provides valuable information on the structure of a system. This information provides decision-makers with an understanding of the key parameters for controlling the system performance and prioritizing parameters. The use of data-driven models and CBML allows for quick calculation of such information. Furthermore, this information provides a plausibility check in terms of engineering by comparing dependencies with domain knowledge.

### *4.3 Rules from local decision trees*

DSE data, local models, and the derivation of rules can provide further information on the system characteristics. Figure 8 shows a tree as a local model for the behavior of different window options of the representative building and provides design rules related to the current configuration to control the heat and radiation passing through the windows. Examining split points allows for the derivation of what-if rules, as examined in [75]. The initial split for the configuration is the size of the windows (Figure 8a), indicating to the decision-maker that small windows require different strategies than large ones. The next two split points (b) identify window orientation as the second most important criterion. Focusing on the south windows following the orange prediction path, the area and g-value, which describe the solar heat gain of the windows, are the key variables for controlling the heat flow at the next levels (c, d). In contrast, east window splitting also includes the u-value, which points to the importance of heat conduction for this orientation. The final prediction (e) shows that orientation and g-value are the guiding variables for the performance of the south windows, indicating that designers should pay attention to these variables at south façades. Furthermore, the upper half of the scatter plots (a, b) indicate that the area and orientation have the highest influence. In particular, increasing area and solar incidence maximizes solar gains and directs designers toward passive solar building designs [76]. Manual studies and extensive sampling of similar climates based on energy simulations performed in other studies confirm the importance of these variables [54], [77].

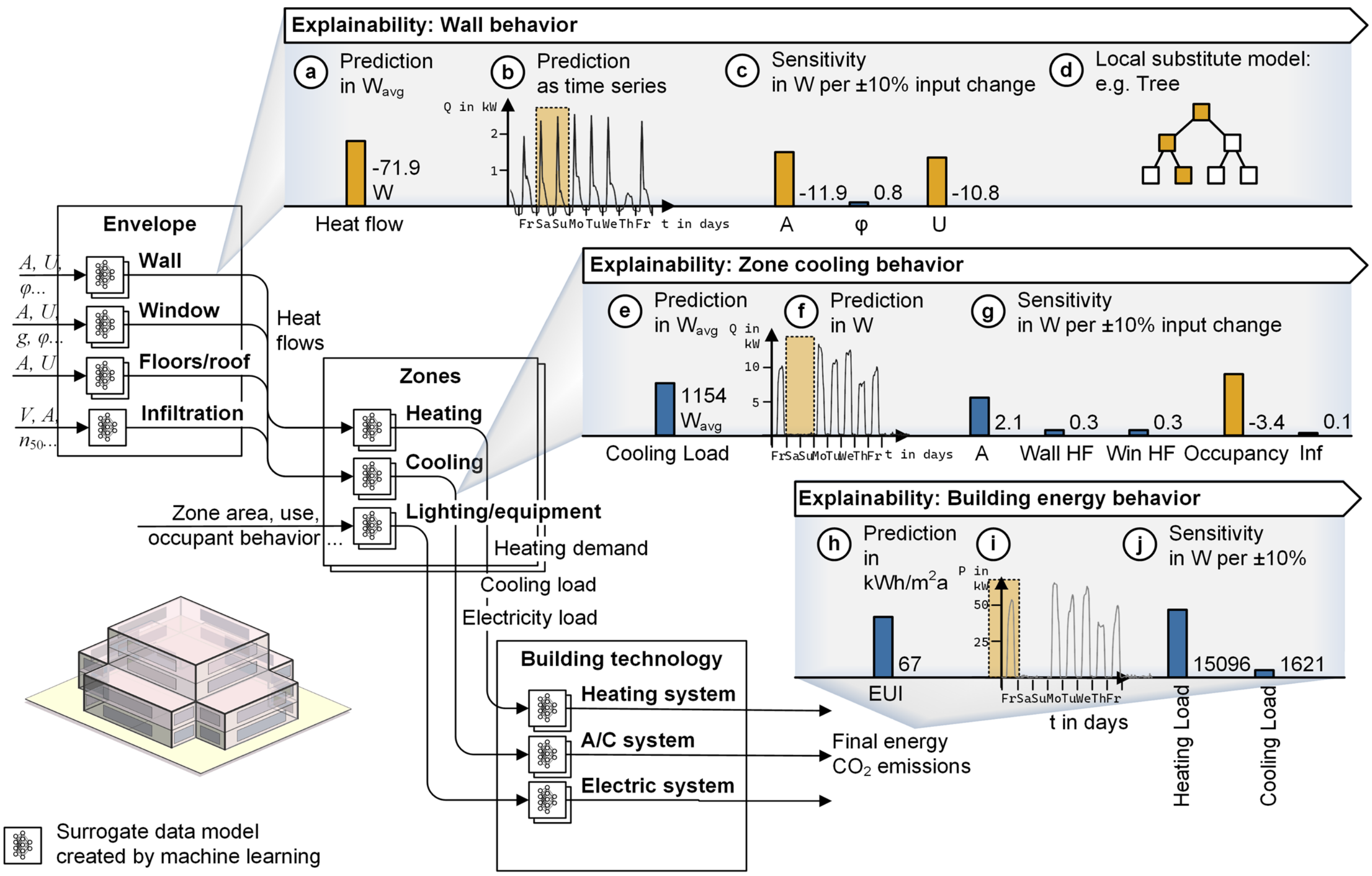

**Fig. 6. Accessing interpretable information at component interfaces enables explainability of predictions of a CBML model.** Accessing activations as engineering quantities with units between components provides numerous engineering insights. The analysis of flows in averages and totals **(a) (e) (h)** and time series (orange dashed box indicates the weekend) **(b) (f) (i);** sensitivities **(c) (g) (j)** and extracted rules based on decision trees **(d)** provide engineering insights. Humans can interpret the results and understand the behavior of the artifacts.

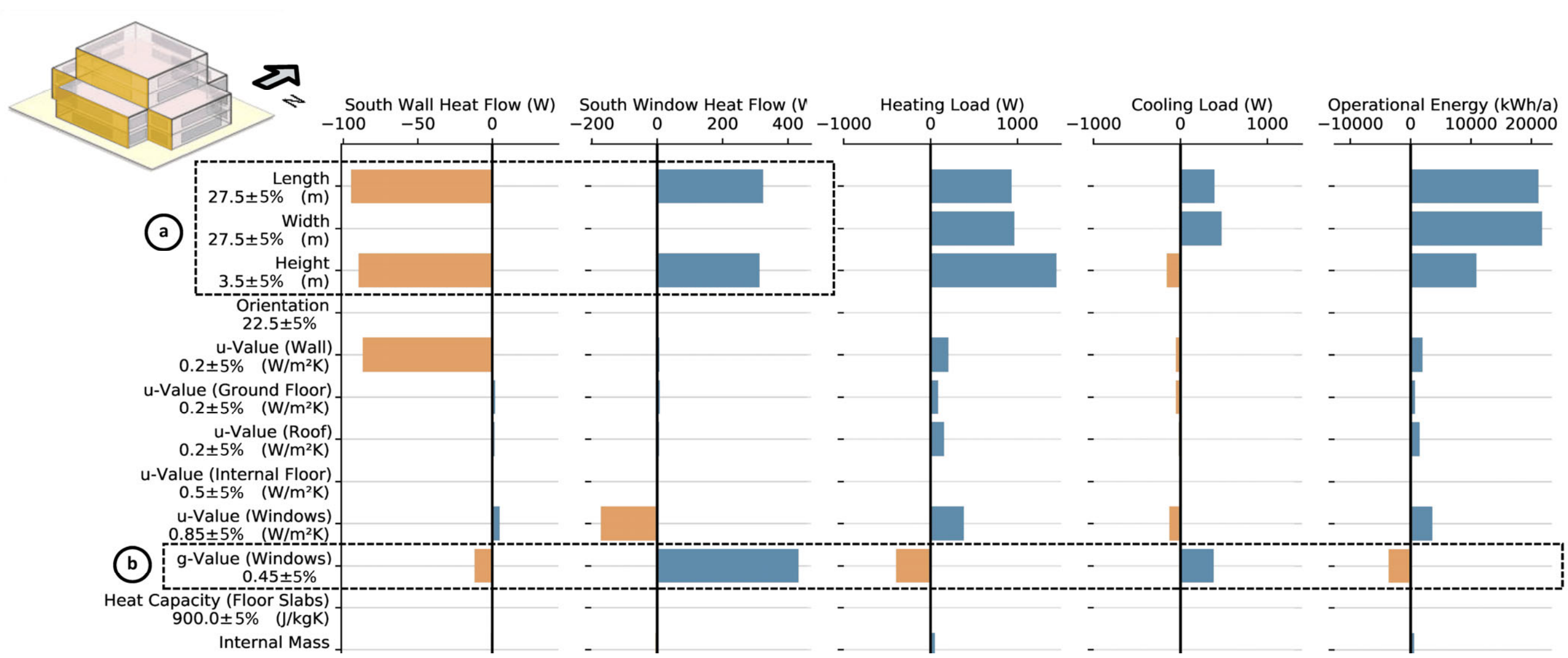

**Fig. 7. Selected sensitivities of the representative test case.** These sensitivities provide domain experts with information to identify the important variables of the design configuration. **(a)** A high sensitivity to geometry and higher gains than losses through the south window are visible. **(b)** The thermal transmittance of the walls and windows (u-value) mainly influences heat losses. **(c)** The examination of the g-value allows us to understand whether gains occur in summer or winter in terms of heating or cooling demand. **(d)** Overall, geometry changes govern the influence on heating and cooling demand as well as total operational energy, followed by the g-value of the windows.

Based on the rules of the tree and underlying data, the application of regression is a method for deriving local engineering equations. For instance, a linear regression for the heat flow through large south windows depending on the area and g-value (Figure 8f) allows decision-makers to not only derive rules from the tree, but also quantitatively assess the effect of changing window size and solar transmittance on the gains and losses.

Second, we tested whether the information derived from the interfaces matched domain knowledge, as described in Section 2.1 and observed in Sections 4.1–4.3. By examining the interfaces for low-energy designs (EUI < 60 kWh/m²a) versus all designs, we derived the statistical distributions at the interfaces, as shown in Figure 11. Interpretation of the distributions in the figure captions corresponds to domain knowledge. In summary, the strategy of large window glazing allowing the sun to enter the building and shading for the summer is related to the classical design strategy called passive solar building design, which is well known in energy-efficient architecture [76]. This demonstrates that the use of direct engineering quantities in the activations at the interfaces connects data-driven modeling to user reasoning, coheres with domain knowledge, and provides explainability. Finally, by covering the entire design space of the representative test case, both tests achieve completeness according to the scheme of Nauta et al. [51]. Finally, the presentation of simple engineering quantities delivers a compactness that matches the user context.

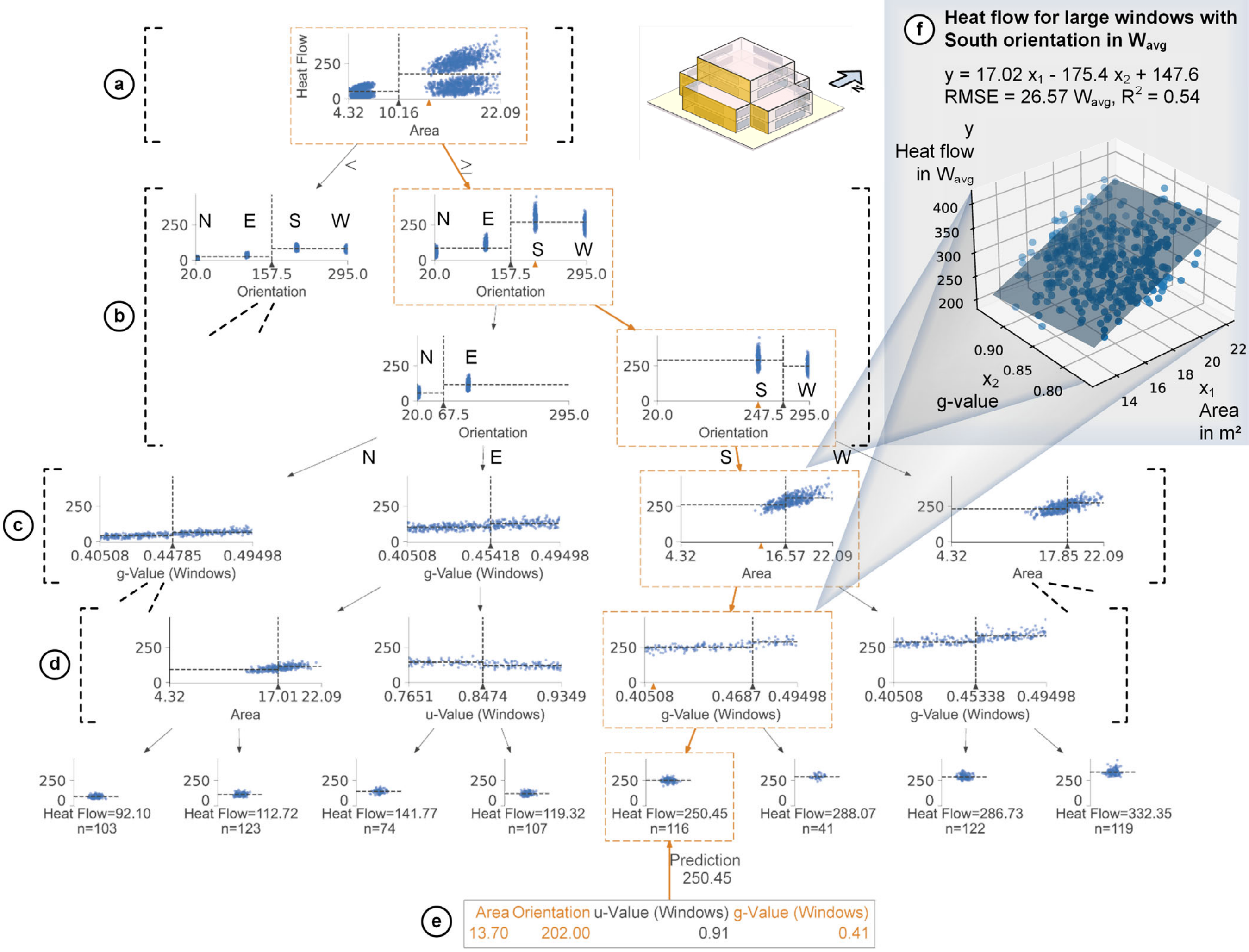

**Fig. 8. Decision tree forming a local surrogate model for the south windows.** Areas **(a)** and **(b)** represent the most important split points. The splitting below **(c, d)** depends on these variables, revealing the specific rules for designing windows, including the u- and g-values. This provides strategic information for the current design **(e)**, for example, delivered as a local surrogate model **(f)**.

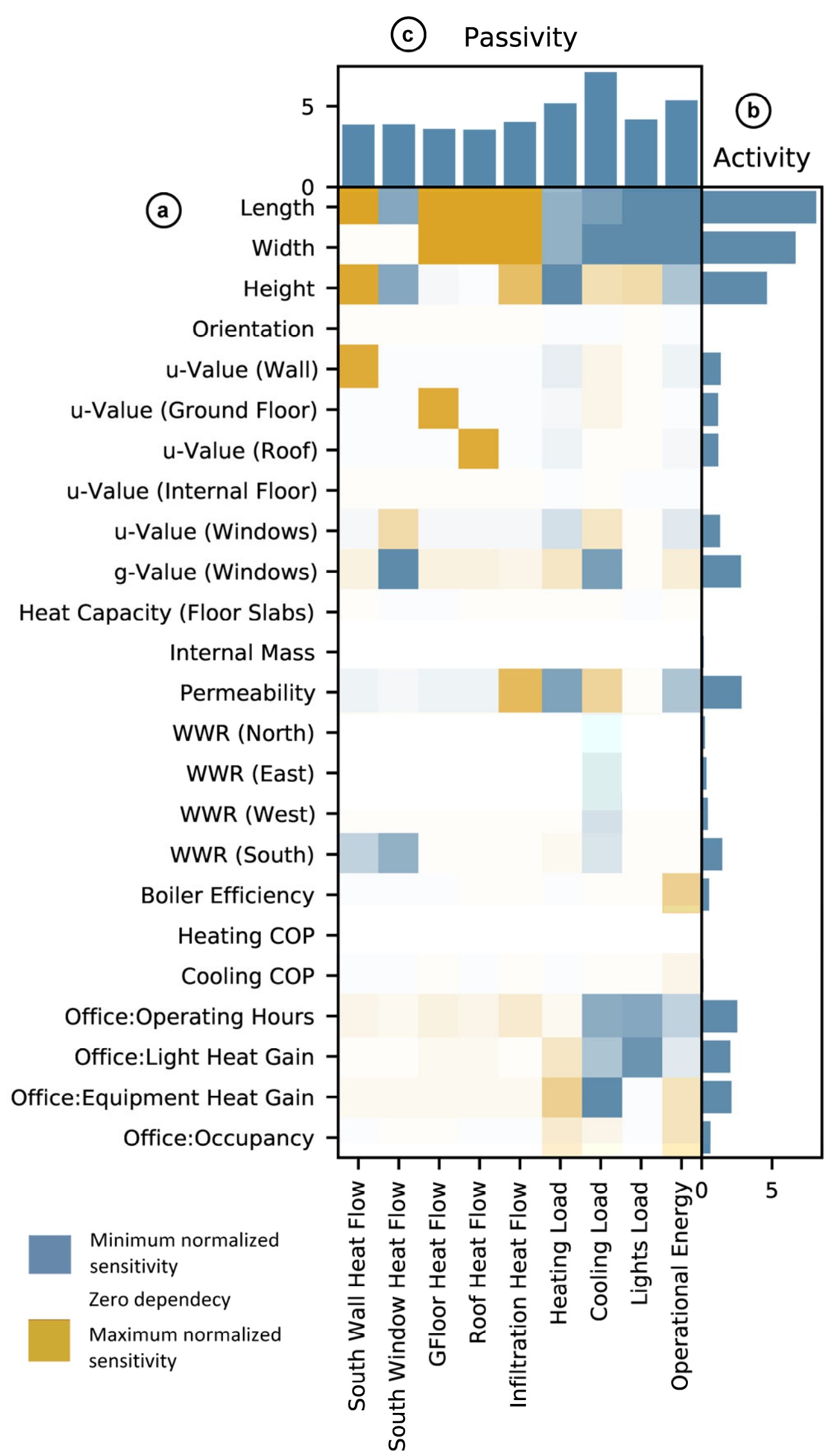

**Fig. 9. The sensitivity matrix shows dependencies between key design variables and internal parameters.** (**a**) Per-column zero-preserving standardized sensitivities based on linear regression coefficients show individual dependencies between design variables and internal parameters. (**b**) The sum of absolute sensitivities determines the extent to which a variable controls the outcome. (**c**) Passivity determines the extent to which a parameter is controlled by the variables.

## *4.4 Evaluation of explainability*

The evaluation contains two tests according to [51]: a white-box test and matching with domain knowledge. First, the white-box test shown in Figure 10 compares the data-driven predictions and physical simulation results of the heat flows and zone loads as internal explanations. A good match with the simulation data proves the reliability of the internal interface information of the data-driven model. Reliability allows designers and engineers to use such numbers, for example, to explain the high or low energy consumption of a design variant and point to the causes of high local demand. Thus, the correctness of the interface information is highly relevant for explainability and trust in the predictions of data-driven models.

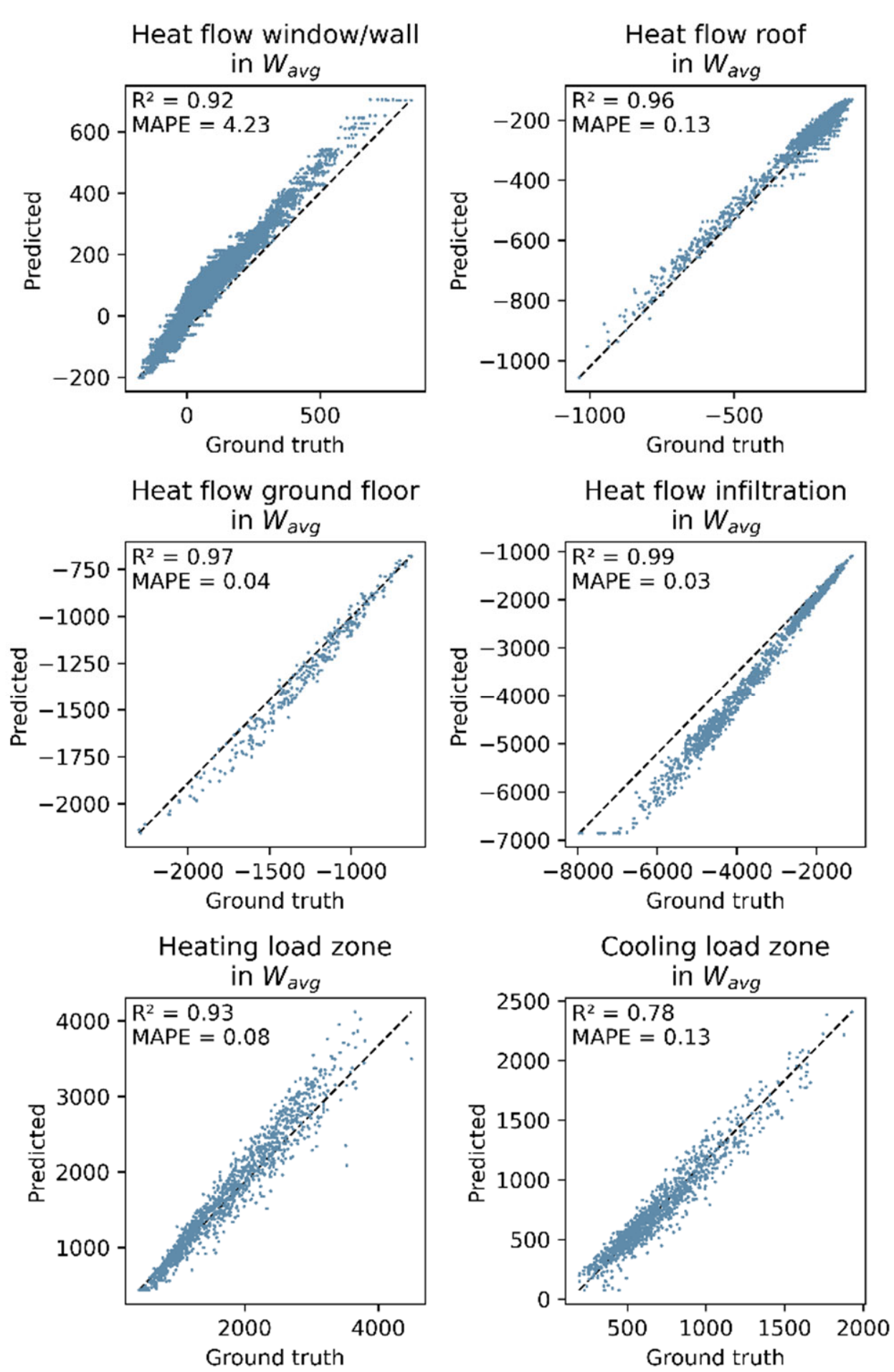

**Fig. 10. White-box test comparing quantities at component interfaces.** The matching of the predicted values with the ground truth at the component interfaces within the component system is a white-box validation.

## 5 Discussion and conclusions

A component-based approach to machine learning offers benefits in systems design and engineering contexts. We demonstrated that CBML provides significantly better generalization than conventional monolithic methods by exploiting domain knowledge in the form of structures. Creating data-driven models following the decomposition schemes of the domain allows for embedding this knowledge and represents a comprehensive range of configurations compared with conventional monolithic data-driven models limited to parametric variation. Recurring components are key to predicting configurations whose structures are not included in the training data; this approach enables a broader prediction context and leads to higher accuracy. In particular, the two different test cases, the randomly created ones and those intentionally equipped with more complexity, demonstrate higher generalization capabilities ($R^2$ of 0.94 instead of 0.71 for the feature-engineered conventional monolithic model).

Additionally, by matching the similarity in the data structure and the probability distribution of the parameters at the component level, as shown in Figure 2, predictions for novel design configurations that differ significantly in structure become possible. In particular, a more complex, manually designed representative case demonstrates that good generalization is possible when adding more complexity to a different design structure. Reliability in extended design space is an advantageous characteristic of the CBML method. The benefit has been proven for the energy-efficient building design domain; evaluation in other domains is part of future research. Based on the similarity of the domain characteristics described in Section 2.2 we expect good transferability, which needs to be proven in future research.

The second advantage of a component-based structure is its potential to explain predictions. A data-driven multi-component model is equivalent to a DNN. The connection of components and their ANN layers with scaling and pooling functions sets up a deep network whose activations are interpretable as engineering quantities at the component interfaces, as shown in Figure 3. This enables a direct understanding of the parameters within the model, and thus, the internal processes of the engineering artifact, as demonstrated for key quantities and flow time series in energy-efficient building design. State-of-the-art approaches of explainability either focus on analyzing input parameters and delivered output, such as the popular LIME and SHAP methods, or they analyze internal activations of data-driven models, which is, however, not reliable and only delivers qualitative information. Compared with this state-of-the-art method, CBML by design offers reliable explainability by interpreting internal DNN activations, and thus, direct quantitative interpretability of the interim steps. For design activities, this direct access is highly relevant for answering what-if questions to understand the behavior of the design artifact and modify its structure. Information derived from local sampling, such as the sensitivities and decision trees, allows us to answer such questions for specific configurations. Sensitivity provides valuable quantitative information regarding parameter changes, directing designers and engineers toward solutions that perform well. Simple trees generated for the local sampling data reveal design rules that align with a specific case. Such rules form a bridge to conventional human design and engineering knowledge, and enhance it by offering quantitative case-related support. The applicability to the case and precise configuration require detailed examination. Hence, the use of data-driven models has a high potential to assist in quantitative reasoning. The component-based approach provides generalization and explainability of such reasoning in engineering design.

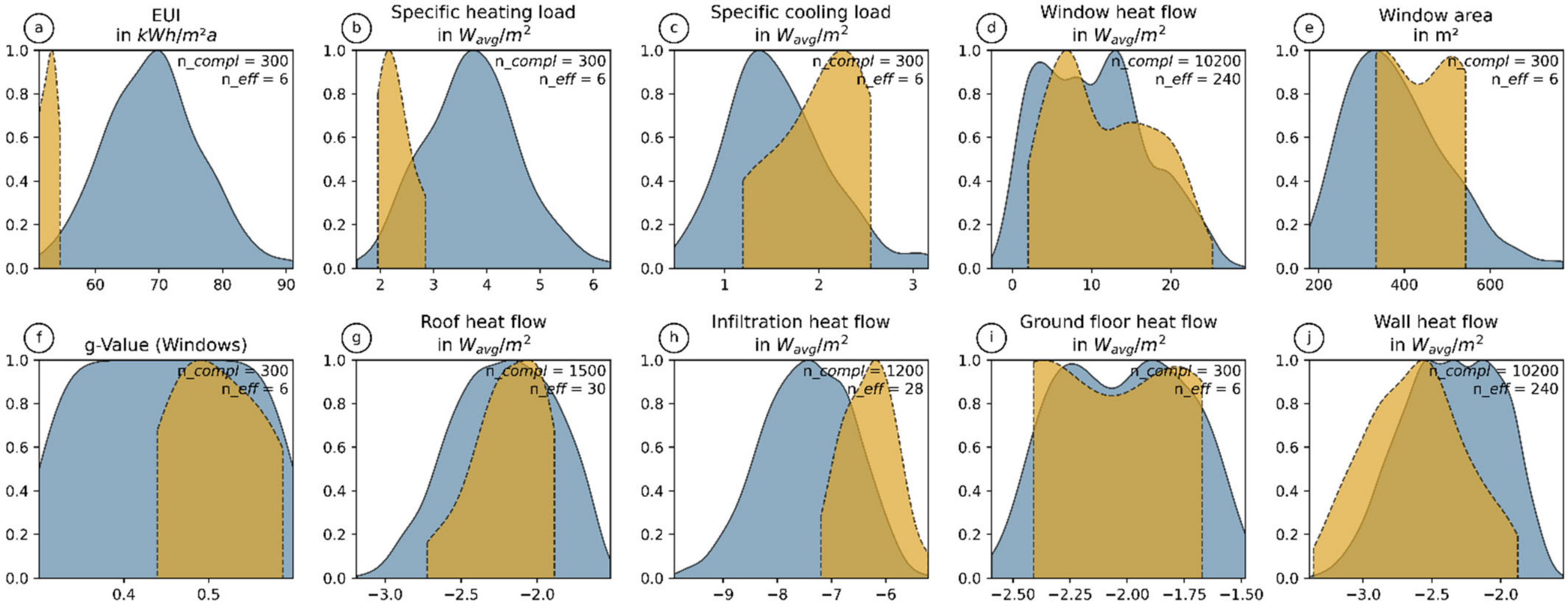

**Fig. 11. Parameter distributions of low-energy designs versus regular designs for the representative case exemplify reasonable engineering interpretability at the component interfaces: (a):** The orange graphs (dashed line) show energy-efficient designs with an energy use intensity (EUI) lower than 55 kWh/m²a, whereas the blue (solid line) shows the complete design space. **(b, c):** Slightly higher cooling loads with more insulation allow for significantly lower heating loads, matching the typical behavior of low-energy buildings. **(d, e, f):** This is achieved by relative high heat gains by larger windows with a high g-value on the one hand. **(g, h):** On the other hand, heat losses of some parts are reduced, such as the roof and infiltration of air through the envelope. **(i):** For ground-floor heat flow, two strategies are visible: insulation, which reduces heat losses in winter, or less insulation, which allows the building to reject heat in summer to reduce cooling loads. **(j):** Wall heat flows increase counterintuitively by reducing insulation. However, this is also a means of rejecting heat in summer, thus reducing cooling loads.

The decomposability determines the applicability of a component-based method. If it is possible to break down a model of an artifact into interacting parts, the application of this method is possible. Such an approach is widespread in engineering design—formalized by the methods of systems engineering—and there is high potential for the application and integration of data-driven models with respective design activities in fields other than energy-efficient building design. Furthermore, it is important for the models to include iterative loops. Dynamic simulations usually solve the thermodynamic behavior of buildings. By using a purely hierarchical structure without any iterations, we demonstrate that it is possible to embed such dynamic behavior in a data-driven model without any iterations. The learning was able to capture the dynamic behavior implicitly, without the need for iteration; for example, behavior caused by the storage of heat in building components that is transferred to indoor spaces. This demonstrates the equivalence of the data-driven models to the iterative solutions of the respective differential equations.

In summary, according to the paradigm of systems engineering, the decomposability of an engineering design using interconnected components characterizes the applicability of component-based data-driven modeling. The advantages of generalization and explainability of the method are vital for data-driven methods to form a surrogate to replace simulations in design and engineering. One the one hand, reduced model effort and fast computation are key to developing intelligent assistance that supports designers and engineers in decision-making with valuable information to achieve more sustainable artifacts. On the other hand, interpretability is a valuable basis for explainable, trustworthy, and responsible applications of machine learning and data-driven methods in engineering design.

## 6 Acknowledgments

The authors acknowledge the support of Deutsche Forschungsgemeinschaft (DFG) for funding this research through grant GE1652/3-1/2 within the researcher unit FOR 2363 and through Heisenberg grant GE1652/4-1. The computational resources and services used in this work were provided by the Flemish Supercomputer Center (VSC), funded by the Research Foundation Flanders (FWO) and the Flemish Government – Department EWI, Belgium.

## 7 Author contributions

**Philipp Geyer:** Conceptualization, Methodology, Software, Validation, Formal analysis, Investigation, Writing - Original draft and revisions, Visualization, Project administration, Funding acquisition

**Manav Mahan Singh:** Methodology, Software, Formal analysis, Investigation, Writing - Review & Editing + Writing Methodology, Visualization

**Xia Chen:** Methodology, Software, Formal analysis, Investigation, Writing - Review & Editing + Writing Methodology, Visualization

## 8 References

[1] K. Amasyali and N. El-Gohary, "A review of data-driven building energy consumption prediction studies," *Renewable and Sustainable Energy Reviews*, vol. 81, pp. 1192–1205, Jan. 2018, doi: 10.1016/J.RSER.2017.04.095.

[2] Y. Wei *et al.*, "A review of data-driven approaches for prediction and classification of building energy consumption," *Renewable and Sustainable Energy Reviews*, vol. 82, pp. 1027–1047, Feb. 2018, doi: 10.1016/j.rser.2017.09.108.

[3] T. Østergård, R. L. Jensen, and S. E. Maagaard, "A comparison of six metamodeling techniques applied to building performance simulations," *Applied Energy*, vol. 211, pp. 89–103, Feb. 2018, doi: 10.1016/j.apenergy.2017.10.102.

[4] L. Zhang *et al.*, "A review of machine learning in building load prediction," *Applied Energy*, vol. 285, p. 116452, Mar. 2021, doi: 10.1016/j.apenergy.2021.116452.

[5] S. Fathi, R. Srinivasan, A. Fenner, and S. Fathi, "Machine learning applications in urban building energy performance forecasting: A systematic review," *Renewable and Sustainable Energy Reviews*, vol. 133, p. 110287, Nov. 2020, doi: 10.1016/j.rser.2020.110287.

[6] P. Westermann and R. Evins, "Surrogate modelling for sustainable building design – A review," *Energy and Buildings*, vol. 198, pp. 170–186, 2019, doi: https://doi.org/10.1016/j.enbuild.2019.05.057.

[7] R. Olu-Ajayi, H. Alaka, I. Sulaimon, F. Sunmola, and S. Ajayi, "Machine learning for energy performance prediction at the design stage of buildings," *Energy for Sustainable Development*, vol. 66, pp. 12–25, Feb. 2022, doi: 10.1016/j.esd.2021.11.002.

[8] H. Salehi and R. Burgueño, "Emerging artificial intelligence methods in structural engineering," *Engineering Structures*, vol. 171, pp. 170–189, 2018, doi: https://doi.org/10.1016/j.engstruct.2018.05.084.

[9] H.-J. Rong, G.-B. Huang, N. Sundararajan, and P. Saratchandran, "Online Sequential Fuzzy Extreme Learning Machine for Function Approximation and Classification Problems," *IEEE Transactions on Systems, Man, and Cybernetics, Part B (Cybernetics)*, vol. 39, no. 4, pp. 1067–1072, Aug. 2009, doi: 10.1109/TSMCB.2008.2010506.

[10] J. Duan, Y. Ou, J. Hu, Z. Wang, S. Jin, and C. Xu, "Fast and Stable Learning of Dynamical Systems Based on Extreme Learning Machine," *IEEE Transactions on Systems, Man, and Cybernetics: Systems*, vol. 49, no. 6, pp. 1175–1185, Jun. 2019, doi: 10.1109/TSMC.2017.2705279.

[11] G. Huang, H. Zhou, X. Ding, and R. Zhang, "Extreme Learning Machine for Regression and Multiclass Classification," *IEEE Transactions on Systems, Man, and Cybernetics, Part B (Cybernetics)*, vol. 42, no. 2, pp. 513–529, 2012, doi: 10.1109/TSMCB.2011.2168604.

[12] S. L. Brunton, B. R. Noack, and P. Koumoutsakos, "Machine Learning for Fluid Mechanics," *Annual Review of Fluid Mechanics*, vol. 52, no. 1, pp. 477–508, 2020, doi: 10.1146/annurev-fluid-010719-060214.

[13] J. H. Lee, J. Shin, and M. J. Realff, "Machine learning: Overview of the recent progresses and implications for the process systems engineering field," *Computers & Chemical Engineering*, vol. 114, pp. 111–121, 2018, doi: https://doi.org/10.1016/j.compchemeng.2017.10.008.

[14] T. Østergård, R. L. Jensen, and S. E. Maagaard, "Early Building Design: Informed decision-making by exploring multidimensional design space using sensitivity analysis," *Energy and Buildings*, vol. 142, pp. 8–22, May 2017, doi: 10.1016/j.enbuild.2017.02.059.

[15] Y. Shen and Y. Pan, "BIM-supported automatic energy performance analysis for green building design using explainable machine learning and multi-objective optimization," *Applied Energy*, vol. 333, p. 120575, Mar. 2023, doi: 10.1016/j.apenergy.2022.120575.

[16] X. Chen and P. Geyer, "Sustainability recommendation system for process-oriented building design alternatives under multi-objective scenarios," presented at the eg-ice 2023, University College London, 2023.

[17] J. P. Alves and J. N. Fidalgo, "Classification of Buildings Energetic Performance Using Artificial Immune Algorithms," in *2019 International Conference on Smart Energy Systems and Technologies (SEST)*, Sep. 2019, pp. 1–6. doi: 10.1109/SEST.2019.8849140.

[18] A. Schlueter, P. Geyer, and S. Cisar, "Analysis of georeferenced building data for the identification and evaluation of thermal microgrids," *Proceedings of the IEEE*, vol. 104, no. 4, 2016, doi: 10.1109/JPROC.2016.2526118.

[19] R. Carli, M. Dotoli, R. Pellegrino, and L. Ranieri, "A Decision Making Technique to Optimize a Buildings' Stock Energy Efficiency," *IEEE Transactions on Systems, Man, and Cybernetics: Systems*, vol. 47, no. 5, pp. 794–807, May 2017, doi: 10.1109/TSMC.2016.2521836.

[20] T. Häring, R. Ahmadiahangar, A. Rosin, and H. Biechl, "Machine Learning Approach for Flexibility Characterisation of Residential Space Heating," in *IECON 2021 – 47th Annual Conference of the IEEE Industrial Electronics Society*, Oct. 2021, pp. 1–6. doi: 10.1109/IECON48115.2021.9589216.

[21] M. Navarro-Cáceres, A. S. Gazafroudi, F. Prieto-Castillo, K. G. Venyagamoorthy, and J. M. Corchado, "Application of artificial immune system to domestic energy management problem," in *2017 IEEE 17th International Conference on Ubiquitous Wireless Broadband (ICUWB)*, Sep. 2017, pp. 1–7. doi: 10.1109/ICUWB.2017.8251010.

[22] G. Montavon, G. Orr, and K.-R. Müller, *Neural Networks: Tricks of the Trade*. Springer, 2012.

[23] J.-S. S. Chou and D.-K. K. Bui, "Modeling heating and cooling loads by artificial intelligence for energy-efficient building design," *Energy and Buildings*, vol. 82, pp. 437–446, Oct. 2014, doi: 10.1016/j.enbuild.2014.07.036.

[24] A. B. Arrieta *et al.*, "Explainable Artificial Intelligence (XAI): Concepts, taxonomies, opportunities and challenges toward responsible AI," *Information Fusion*, vol. 58, pp. 82–115, 2020, doi: https://doi.org/10.1016/j.inffus.2019.12.012.

[25] A. Holzinger, A. Saranti, C. Molnar, P. Biecek, and W. Samek, "Explainable AI Methods - A Brief Overview," in *xxAI - Beyond Explainable AI: International Workshop, Held in Conjunction with ICML 2020, July 18, 2020, Vienna, Austria, Revised and Extended Papers*, A. Holzinger, R. Goebel, R. Fong, T. Moon, K.-R. Müller, and W. Samek, Eds., Cham: Springer International Publishing, 2022, pp. 13–38. doi: 10.1007/978-3-031-04083-2_2.

[26] L. Longo *et al.*, "Explainable Artificial Intelligence (XAI) 2.0: A Manifesto of Open Challenges and Interdisciplinary Research Directions," Oct. 30, 2023, *arXiv*: arXiv:2310.19775. doi: 10.48550/arXiv.2310.19775.

[27] R. Roscher, B. Bohn, M. F. Duarte, and J. Garcke, "Explainable Machine Learning for Scientific Insights and Discoveries," *IEEE Access*, vol. 8, pp. 42200–42216, 2020, doi: 10.1109/ACCESS.2020.2976199.

[28] W. J. Murdoch, C. Singh, K. Kumbier, R. Abbasi-Asl, and B. Yu, "Definitions, methods, and applications in interpretable machine learning," *Proc Natl Acad Sci U S A*, vol. 116, no. 44, pp. 22071–22080, Oct. 2019, doi: 10.1073/pnas.1900654116.

[29] C. Molnar, G. Casalicchio, and B. Bischl, "Interpretable Machine Learning – A Brief History, State-of-the-Art and Challenges," in *ECML PKDD 2020 Workshops*, I. Koprinska, M. Kamp, A. Appice, C. Loglisci, L. Antonie, A. Zimmermann, R. Guidotti, Ö. Özgöbek, R. P. Ribeiro, R. Gavaldà, J. Gama, L. Adilova, Y. Krishnamurthy, P. M. Ferreira, D. Malerba, I. Medeiros, M. Ceci, G. Manco, E. Masciari, Z. W. Ras, P. Christen, E. Ntoutsi, E. Schubert, A. Zimek, A. Monreale, P. Biecek, S. Rinzivillo, B. Kille, A. Lommatzsch, and J. A. Gulla, Eds., Cham: Springer International Publishing, 2020, pp. 417–431. doi: 10.1007/978-3-030-65965-3_28.

[30] M. Manfren, P. AB. James, and L. Tronchin, "Data-driven building energy modelling – An analysis of the potential for generalisation through interpretable machine learning," *Renewable and Sustainable Energy Reviews*, vol. 167, p. 112686, Oct. 2022, doi: 10.1016/j.rser.2022.112686.

[31] P. Geyer and A. Schlüter, "Automated metamodel generation for Design Space Exploration and decision-making – A novel method supporting performance-oriented building design and retrofitting," *Applied Energy*, vol. 119, pp. 537–556, Apr. 2014, doi: 10.1016/j.apenergy.2013.12.064.

[32] Z. Chen, F. Xiao, F. Guo, and J. Yan, "Interpretable machine learning for building energy management: A state-of-the-art review," *Advances in Applied Energy*, vol. 9, p. 100123, Feb. 2023, doi: 10.1016/j.adapen.2023.100123.

[33] S. Shams Amiri, S. Mottahedi, E. R. Lee, and S. Hoque, "Peeking inside the black-box: Explainable machine learning applied to household transportation energy consumption," *Computers, Environment and Urban Systems*, vol. 88, p. 101647, Jul. 2021, doi: 10.1016/j.compenvurbsys.2021.101647.

[34] P. Cortez and M. J. Embrechts, "Opening black box Data Mining models using Sensitivity Analysis," in *2011 IEEE Symposium on Computational Intelligence and Data Mining (CIDM)*, Apr. 2011, pp. 341–348. doi: 10.1109/CIDM.2011.5949423.

[35] S. Singaravel, J. Suykens, H. Janssen, and P. Geyer, "Explainable deep convolutional learning for intuitive model development by non–machine learning domain experts," *Design Science*, vol. 6, p. e23, 2020, doi: 10.1017/dsj.2020.22.

[36] X. Chen and P. Geyer, "Machine assistance in energy-efficient building design: A predictive framework toward dynamic interaction with human decision-making under uncertainty," *Applied Energy*, vol. 307, p. 118240, Feb. 2022, doi: 10.1016/j.apenergy.2021.118240.

[37] J.-Y. Kim and S.-B. Cho, "Electric Energy Consumption Prediction by Deep Learning with State Explainable Autoencoder," *Energies*, vol. 12, no. 4, Art. no. 4, Jan. 2019, doi: 10.3390/en12040739.

[38] S.-I. Ao, B. B. Rieger, and M. Amouzegar, *Machine learning and systems engineering*, vol. 68. Springer Science & Business Media, 2010.

[39] Y. S. Kim and C. S. Park, "Real-time predictive control of HVAC systems for factory building using lightweight data-driven model," *Journal of Building Performance Simulation*, vol. 16, no. 5, pp. 507–525, Sep. 2023, doi: 10.1080/19401493.2023.2182363.

[40] J. H. Lee, J. Shin, and M. J. Realff, "Machine learning: Overview of the recent progresses and implications for the process systems engineering field," *Computers & Chemical Engineering*, vol. 114, pp. 111–121, Jun. 2018, doi: 10.1016/j.compchemeng.2017.10.008.

[41] A. Kumbhar, P. G. Dhawale, S. Kumbhar, U. Patil, and P. Magdum, "A comprehensive review: Machine learning and its application in integrated power system," *Energy Reports*,

vol. 7, pp. 5467–5474, Nov. 2021, doi: 10.1016/j.egyr.2021.08.133.

[42] "Brick Schema." Accessed: Apr. 26, 2024. [Online]. Available: https://brickschema.org/

[43] J. Shi, Z. Pan, L. Jiang, and X. Zhai, "An ontology-based methodology to establish city information model of digital twin city by merging BIM, GIS and IoT," *Advanced Engineering Informatics*, vol. 57, p. 102114, Aug. 2023, doi: 10.1016/j.aei.2023.102114.

[44] N. M. Tomašević, M. Č. Batić, L. M. Blanes, M. M. Keane, and S. Vraneš, "Ontology-based facility data model for energy management," *Advanced Engineering Informatics*, vol. 29, no. 4, pp. 971–984, Oct. 2015, doi: 10.1016/j.aei.2015.09.003.

[45] D. Wolosiuk and A. Mahdavi, "Application of ontologically streamlined data for building performance analysis," in *ECPPM 2021 - eWork and eBusiness in Architecture, Engineering and Construction*, CRC Press, 2021.

[46] P. Zhou and N. El-Gohary, "Semantic information alignment of BIMs to computer-interpretable regulations using ontologies and deep learning," *Advanced Engineering Informatics*, vol. 48, p. 101239, Apr. 2021, doi: 10.1016/j.aei.2020.101239.

[47] P. Delgoshaei, M. Heidarinejad, and M. A. Austin, "Combined ontology-driven and machine learning approach to monitoring of building energy consumption," in *2018 Building Performance Modeling Conference and SimBuild, Chicago, IL*, 2018, pp. 667–674. Accessed: Apr. 26, 2024. [Online]. Available: https://publications.ibpsa.org/proceedings/simbuild/2018/papers/simbuild2018_C092.pdf

[48] S. J. Pan and Q. Yang, "A Survey on Transfer Learning," *IEEE Transactions on Knowledge and Data Engineering*, vol. 22, no. 10, pp. 1345–1359, 2010, doi: 10.1109/TKDE.2009.191.

[49] K. Weiss, T. M. Khoshgoftaar, and D. Wang, "A survey of transfer learning," *Journal of Big Data*, vol. 3, no. 1, p. 9, 2016, doi: 10.1186/s40537-016-0043-6.

[50] R. R. Hoffman, S. T. Mueller, G. Klein, and J. Litman, "Metrics for Explainable AI: Challenges and Prospects," Feb. 01, 2019, *arXiv*: arXiv:1812.04608. doi: 10.48550/arXiv.1812.04608.

[51] M. Nauta *et al.*, "From Anecdotal Evidence to Quantitative Evaluation Methods: A Systematic Review on Evaluating Explainable AI," May 31, 2022, *arXiv*: arXiv:2201.08164. doi: 10.48550/arXiv.2201.08164.

[52] P. Geyer and S. Singaravel, "Component-based machine learning for performance prediction in building design," *Applied Energy*, vol. 228, pp. 1439–1453, Oct. 2018, doi: 10.1016/j.apenergy.2018.07.011.

[53] J. Abualdenien *et al.*, "Consistent management and evaluation of building models in the early design stages," *Journal of Information Technology in Construction (ITcon)*, vol. 25, no. 13, pp. 212–232, Mar. 2020, doi: 10.36680/j.itcon.2020.013.

[54] M. M. Singh and P. Geyer, "Information requirements for multi-level-of-development BIM using sensitivity analysis for energy performance," *Advanced Engineering Informatics*, vol. 43, p. 101026, 2020, doi: https://doi.org/10.1016/j.aei.2019.101026.

[55] X. Chen, M. M. Singh, and P. Geyer, "Utilizing domain knowledge: Robust machine learning for building energy performance prediction with small, inconsistent datasets," *Knowledge-Based Systems*, vol. 294, p. 111774, Jun. 2024, doi: 10.1016/j.knosys.2024.111774.

[56] R. Haberfellner, O. de Weck, E. Fricke, and S. Vössner, "Systems Engineering: Fundamentals and Applications," 2019.

[57] M. Kreimeyer and U. Lindemann, *Complexity metrics in engineering design*. Berlin [u.a.]: Springer, 2011. [Online]. Available: http://d-nb.info/1010922327/04

[58] X. Chen and P. Geyer, "Machine assistance: A predictive framework toward dynamic interaction with human decision-making under uncertainty in energy-efficient building design," *submitted*, 2021.

[59] S. Singaravel, J. Suykens, and P. Geyer, "Deep-learning neural-network architectures and methods: Using component-based models in building-design energy prediction," *Advanced Engineering Informatics*, vol. 38, pp. 81–90, Oct. 2018, doi: 10.1016/j.aei.2018.06.004.

[60] R. Sacks, C. Eastman, G. Lee, and P. Teicholz, *BIM handbook: a guide to building information modeling for owners, designers, engineers, contractors, and facility managers*. John Wiley & Sons, 2018.

[61] W. Samek and K.-R. Müller, "Towards Explainable Artificial Intelligence," in *Explainable AI: Interpreting, Explaining and Visualizing Deep Learning*, W. Samek, G. Montavon, A. Vedaldi, L. K. Hansen, and K.-R. Müller, Eds., in Lecture Notes in Computer Science. , Cham: Springer International Publishing, 2019, pp. 5–22. doi: 10.1007/978-3-030-28954-6_1.

[62] S. Bach, A. Binder, G. Montavon, F. Klauschen, K.-R. Müller, and W. Samek, "On Pixel-Wise Explanations for Non-Linear Classifier Decisions by Layer-Wise Relevance Propagation," *PLOS ONE*, vol. 10, no. 7, pp. 1–46, 2015, doi: 10.1371/journal.pone.0130140.

[63] M. T. Ribeiro, S. Singh, and C. Guestrin, "'Why Should I Trust You?': Explaining the Predictions of Any Classifier," in *Proceedings of the 22Nd ACM SIGKDD International Conference on Knowledge Discovery and Data Mining*, in KDD '16. New York, NY, USA: ACM, 2016, pp. 1135–1144. doi: 10.1145/2939672.2939778.

[64] M. T. Ribeiro, S. Singh, and C. Guestrin, "Model-Agnostic Interpretability of Machine Learning," in *ICML Workshop on Human Interpretability in Machine Learning*, New York, Jun. 2016.

[65] K. Menberg, Y. Heo, and R. Choudhary, "Sensitivity analysis methods for building energy models: Comparing computational costs and extractable information," *Energy and Buildings*, vol. 133, pp. 433–445, Dec. 2016, doi: 10.1016/j.enbuild.2016.10.005.

[66] S. Mohseni, N. Zarei, and E. D. Ragan, "A Multidisciplinary Survey and Framework for Design and Evaluation of Explainable AI Systems," Aug. 05, 2020, *arXiv*: arXiv:1811.11839. doi: 10.48550/arXiv.1811.11839.

[67] L. von Rueden *et al.*, "Informed Machine Learning – A Taxonomy and Survey of Integrating Knowledge into Learning Systems," *IEEE Trans. Knowl. Data Eng.*, pp. 1–1, 2021, doi: 10.1109/TKDE.2021.3079836.

[68] K. Beckh *et al.*, "Explainable Machine Learning with Prior Knowledge: An Overview," *arXiv:2105.10172 [cs]*, May 2021, Accessed: Dec. 29, 2021. [Online]. Available: http://arxiv.org/abs/2105.10172

[69] J. Clarke, *Energy Simulation in Building Design*. Oxford: Butterworth-Heinemann, 2001. [Online]. Available: http://www.sciencedirect.com/science/article/pii/B9780750650823500012

[70] U.S. Department of Energy's (DOE), "EnergyPlus." Accessed: Jun. 01, 2021. [Online]. Available: https://energyplus.net/

[71] F. N. Najm, *Circuit Simulation*. John Wiley & Sons, 2010.

[72] J. G. de Jalon and E. Bayo, *Kinematic and Dynamic Simulation of Multibody Systems: The Real-Time Challenge*. Springer Science & Business Media, 2012.

[73] R. Andújar, J. Roset, and V. Kilar, “Interdisciplinary approach to numerical methods for structural dynamics,” *World applied sciences journal*, vol. 14, no. 8, pp. 1046–1053, 2011.
[74] S. Eppinger and T. Browning, *Design structure matrix methods and applications*. Cambridge, Mass: MIT Press, 2012.
[75] X. Chen, J. Abualdenien, M. M. Singh, A. Borrmann, and P. Geyer, “Introducing causal inference in the energy-efficient building design process,” *Energy and Buildings*, vol. 277, p. 112583, Dec. 2022, doi: 10.1016/j.enbuild.2022.112583.
[76] J. D. Balcomb, *Passive Solar Buildings*. MIT Press, 1992.
[77] T. Potrč Obrecht, M. Premrov, and V. Žegarac Leskovar, “Influence of the orientation on the optimal glazing size for passive houses in different European climates (for non-cardinal directions),” *Solar Energy*, vol. 189, pp. 15–25, Sep. 2019, doi: 10.1016/j.solener.2019.07.037.
[78] M. M. Singh, “Validation of Early Design Stage EnergyPlus Model for Office Building (one-zone-per-floor model),” Sep. 2021, doi: 10.17632/5kskbt6w2s.1.
[79] M. M. Singh, S. Singaravel, R. Klein, and P. Geyer, “Quick energy prediction and comparison of options at the early design stage,” *Advanced Engineering Informatics*, vol. 46, p. 101185, Oct. 2020, doi: 10.1016/j.aei.2020.101185.
[80] Passivhaus Institut, *Passive House requirements*. 2015.
[81] Bundesministerium für Wirtschaft und Energie, “Energy Saving Ordinance (EnEV) 2014, Zweite Verordnung zur Änderung der Energieeinsparverordnung, BGBl I Seite 3951 vom 18.Nov.2013.” Accessed: May 21, 2021. [Online]. Available: https://www.bmwi.de/Redaktion/DE/Downloads/Gesetz/zweite-verordnung-zur%20aenderung-der-energieeinsparverordnung.html
[82] X. Chen, M. M. Singh, and P. Geyer, “Component-based machine learning for predicting representative time-series of energy performance in building design,” presented at the 28th International Workshop on Intelligent Computing in Engineering, Berlin, 2021.
[83] Y. Freund, R. Schapire, and N. Abe, “A short introduction to boosting,” *Journal-Japanese Society For Artificial Intelligence*, vol. 14, no. 771–780, p. 1612, 1999.
[84] Guolin Ke *et al.*, “Lightgbm: A highly efficient gradient boosting decision tree,” *Advances in neural information processing systems*, vol. 30, pp. 3146–3154, 2017.
[85] Guolin Ke and et al., *microsoft/LightGBM*. (May 24, 2021). C++. Accessed: May 24, 2021. [Online]. Available: https://github.com/microsoft/LightGBM
[86] L. Breiman, J. Friedman, C. J. Stone, and R. A. Olshen, *Classification and regression trees*. CRC press, 1984.
[87] B. Hssina, A. Merbouha, H. Ezzikouri, and M. Erritali, “A comparative study of decision tree ID3 and C4. 5,” *International Journal of Advanced Computer Science and Applications*, vol. 4, no. 2, pp. 13–19, 2014.
[88] F. Pedregosa *et al.*, “Scikit-learn: Machine learning in Python,” *the Journal of machine Learning research*, vol. 12, pp. 2825–2830, 2011.
[89] P Terence Parr, Tudor Lapusan, and Prince Grover, “GitHub - parrt/dtreeviz: A python library for decision tree visualization and model interpretation.” Accessed: May 24, 2021. [Online]. Available: https://github.com/parrt/dtreeviz

# 9 Appendix

## *9.1 Appendix A: Data Generation Using Dynamic Energy Simulation*

ML model training and testing requires a large amount of training data covering different design configurations. As it is difficult to collect such examples from real buildings, a common approach is to develop and validate a dynamic simulation model and to use it to generate synthetic data. We developed an EnergyPlus (EP) [70] simulation model for an existing office building in Munich and validated this model against data measured for two years. The building’s parameters are listed in Table 1 and floor plan is shown in Figure 12. The simulation model as been validated by comparison with the measurement from an existing building that is covered by the data: The measured total of heating and cooling energy demand is 43.97 MWh/a whereas the simulated value is 43.98 MWh/a. The simulated lighting energy demand is 21 MWh/a, for which the real data is not available. The total energy demand corresponds to 54.6 kWh/m$^2$a. The simulation model and measured data is available on Mendeley datasets [78].

Representative key design parameters and their ranges have been selected to generate data covering design configurations. In this selection previous studies and relevant German standards served as reference [79]–[81]. The parameters used in this article are shown in Table 2. These parameters are selected based on their relevance for the design activity at the early design stage as known from previous examination [54].

The design parameters are sampled using Sobol scheme to generate 1000 random samples for training data. For each sample, an EP model is created and simulated. Simulation results collected as training data include average and totals as well as time-series for heat flows, loads, and energy consumption. The training data consists of box-shaped building samples (Fig. 1a), while the test dataset consists of random shapes (Fig. 1b, Test Dataset 1) and a manually designed representative shape (Fig. 1b, Test Dataset 2). Latin hypercube sampling is used to generate 300 random samples for the test dataset. Additional 300 samples are generated for the representative case. The design cases of the test and representative dataset are more complex than the training dataset. This allows for testing the generalizability of CBML approach on complex design cases.

**Table 1 Parameters of the simulation model for the real building**

| Parameters | Unit | Value |
|---|---|---|
| Floor-to-floor Height | M | 3.25 |
| Number of floors | - | 3 |
| u-value (Wall) | | 0.18 |
| u-value (Ground Floor) | $W/m^2K$ | 0.19 |
| u-value (Roof) | | 0.15 |
| u-value (Window) | | 0.87 |
| g-value | | 0.35 |
| Heat Capacity (Slab) | J/kgK | 800 |
| Permeability | $m^3/m^2h$ | 6 |
| Internal Mass | $kJ/m^2K$ | 120 |
| Operating Hours | H | 11 |
| Occupant Load | $m^2$/Person | 24 |
| Light Heat Load | $W/m^2$ | 6 |
| Equipment Heat Load | | 12 |
| Heating COP | | 2.8 |
| Cooling COP | - | 3.6 |
| Boiler Efficiency | | 0.95 |

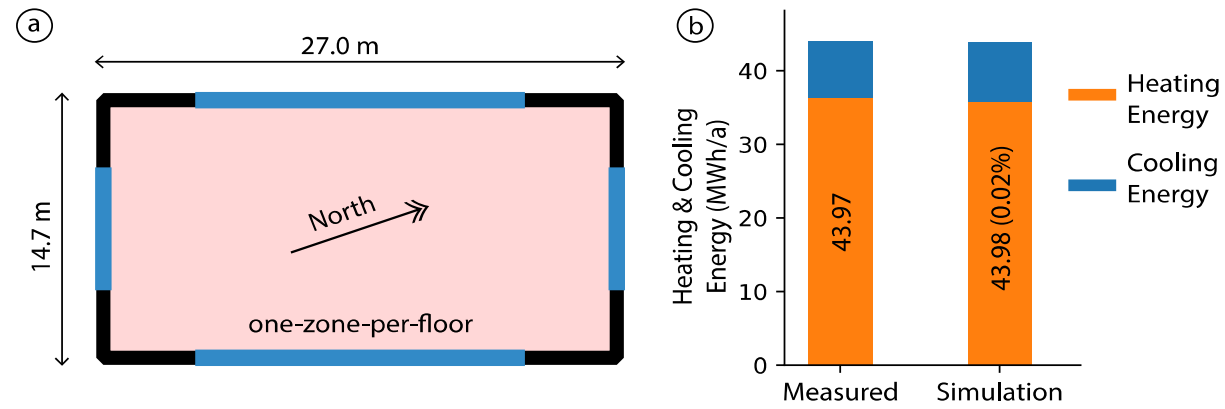

**Fig. 12.Comparison of simulated and actual energy consumption. (a)** building floor plan (zoning model). **(b)** measured and simulated heating & cooling energy requirements

### 9.2 *Appendix B: Monolithic Baseline Experiment*

Previously, researchers have proposed monolithic approaches to develop ML-based energy prediction model for different building shapes. This approach uses the parametric building characteristics such as area, u-values, window-to-wall ratios (WWR) etc. and shape factor or relative compactness to represent the shape characteristics. This additional feature represents the volume to façade surface ratio of the building. Using this approach, this study develops a monolithic ML model as baseline model representing best practice of data-driven modeling. The details of input features are provided in Table 3.

**Table 2 Parameters and their ranges for training and test datasets**

| Parameter | Unit | Min | Max |
|---|---|---|---|
| Length/Width [1] | m | 12 | 30 |
| Ground Floor Area [2] | $m^2$ | 250 | 800 |
| Height | m | 3 | 4 |
| Orientation | Degrees | 0 | 90 |
| Number of Floors | - | 2 | 5 |
| u-value (Wall) | | 0.15 | 0.25 |
| u-value (Ground Floor) | | 0.15 | 0.25 |
| u-value (Roof) | $W/m^2K$ | 0.15 | 0.25 |
| u-value (Internal Floor) | | 0.4 | 0.6 |
| u-value (Windows) | | 0.7 | 1.0 |
| g-Value (Windows) | - | 0.3 | 0.6 |
| Heat Capacity (Slab) | $J/m^3K$ | 800 | 1000 |
| Internal Mass Heat Capacity | $kJ/m^2K$ | 60 | 120 |
| Permeability | $m^3/m^2h$ | 6 | 9 |
| WWR [3] | - | 0.1 | 0.5 |
| Boiler Efficiency | | 0.92 | 0.98 |
| Heating COP | - | 2.5 | 4.5 |
| Cooling COP | | 2.5 | 4.5 |
| Operating Hours | h | 10 | 12 |
| Light Heat Gain | $W/m^2$ | 6 | 10 |
| Equipment Heat Gain | | 10 | 14 |
| Occupancy | Person/$m^2$ | 16 | 24 |

[1] Length and Width box-shaped and representative building cases
[2] Ground Floor Area for random shapes buildings
[3] Window-to-wall ratio (WWR) varies independently in each direction

The baseline monolithic ML model is trained as simple artificial neural network with one input layer, one hidden layer, and one output layer. This study uses L2 regularization and early stopping to prevent overfitting. 20% of the training samples have been used to tune hyperparameters. After a few initial runs, the learning rate has been fixed 0.001, the batch size as one-fifth of the training dataset, activation function as rectified linear unit. A total of sixteen combinations of hyperparameters (four values for number of neurons and four values for the value of regularization coefficient) have been tried and the model with the least validation loss has been kept for further research. Table 3 shows the details of hyperparameters, used for training the monolithic ML model.

**Table 3 Input and output of monolithic ML model serving as best-practice baseline**

| Input | | Output |
|---|---|---|
| Floor Area (m$^2$)<br>Height (m)<br>Number of Floors (-)<br>Relative Compactness (m$^3$/m$^2$)<br>u-value [Wall, Ground Floor, Roof, Windows] (W/m$^2$K)<br>g-Value (-)<br>Permeability (m$^3$/m$^2$h)<br>Internal Mass (J/m$^2$K) | WWR [North, East, West, South] (-)<br>Operating Hours (h)<br>Light Heat Gain (W/m$^2$)<br>Equipment Heat Gain (W/m$^2$)<br>Occupancy (Person/m$^2$)<br>Setpoint [Heating Cooling] (℃)<br>Boiler Efficiency (-)<br>Coefficient of Performance [Heating Cooling] (-) | Annual Energy Demand (kWh/a) |

**Table 4 Details of hyperparameters used for training ML components**

| Hyperparameter | Values |
|---|---|
| Number of Neurons | 200, 400, 600, 800 |
| Regularization Coefficient | 0.0003, 0.0001, 0.00003, 0.00001 |
| Learning Rate | 0.001 |
| Batch Size | One-fifth of sample size |
| Activation | Rectified Linear Unit (ReLU) |

### *9.3 Appendix C: Component-Model Generation*

In this approach of CBML, nine ML components are arranged in hierarchical order to predict building energy demand by use of ML for regression. The first level contains five ML components that predict heat flows, corresponding to elements and properties of the building envelope, i.e., wall, window, floor, roof, and infiltration. The second level contains three ML components that predict zone loads related to heating, cooling, and lighting. Finally, the third level has one ML component related to building systems and their properties (heating, cooling and electric) to predict building final energy demand. The input for each ML component is mentioned in Table 5.

Each ML component has a typical artificial neural network (ANN) architecture. There is one input layer, one hidden layer, and one output layer. During component training, 20% of the training data is kept as validation dataset. After few trial runs, the learning rate has been fixed to 0.001, batch size to one-fifth of the training dataset, and activation function to rectified linear unit (ReLU). The model uses both L2 regularization and early stopping to prevent overfitting. Sixteen different combinations of coefficients for regularization and the number of neurons in the hidden layer have been tested. The best model with the least validation error has been retained for further research.

**Table 5 Input and output of the ML components**

| ML Component | Input | Output |
|---|---|---|
| Wall | Area (m$^2$), orientation (°), u-value (W/m$^2$K) | Heat Flow (W) |
| Window | Area (m$^2$), orientation (°), u-value (W/m$^2$K), g-value (-) | |
| Floor/ roof | Area (m$^2$), u-value (W/m$^2$K), heat capacity (J/kgK) | |
| Infiltration | Area (m$^2$), height (m), permeability (m$^3$/m$^2$h), heat capacity (J/kgK) | |
| Zone heating/ cooling load | Area (m$^2$), [wall/window/floor/ roof/infiltration], heat flow (W), Internal Mass Heat Capacity (kJ/m$^2$K), [light/ equipment], heat gain (W/m $^2$), operating hours (h), occupancy (Person/m2) | Heating/ cooling Load (W) |
| Zone lighting load | Area (m$^2$), light heat gain (W/m$^2$), operating hours (h), window area (m$^2$), g-value (-) | Lighting Load (W) |
| Building energy demand | Boiler efficiency (-), [heating/ cooling] COP (-), [heating/ cooling/ lighting] load (W/m$^2$) | Annual Energy Demand (kWh/a) |

### *9.4 Appendix D: Component-Model Generation for Time-Series Predictions*

For time series predictors, a component also following the previously used ML-for-regression scheme has been trained and tested with an altered input/output data structure in an additional study [82]. In the data generation process, heat flow, load and energy consumption include the extra dimension time. Furthermore, time series information on the climate is included in training. Feature engineering served to extract and strengthen the periodic characteristics for the model: In our approach, timestamp formatting (year, month, week, day, day of the week, week of the month, hour, etc. plus Boolean value "is weekday") has been applied. Another important aspect for the time series prediction is autocorrelation. Feature engineering techniques for shifting, lagging, and window averaging are usually combined in time series data related regression or prediction tasks. In practice, for autocorrelation, input features from *n* previous states have been used in training phase. Depending on different periodic characteristics, *n* is set to 3 to 7 days or 12 to 24 hours.

Essentially, the time-series predictor is a regression model, too. The regression algorithm itself has no special requirement on tailoring to fit time series. For the balance of accuracy, difficulty of implementation and interpretability, we used the ensemble method Gradient Boosting Decision Tree (GBDT), which is developed following the concept of boosting [83]. In our implementation, LightGBM [84] using an open-source python library implementation [85] is chosen to fit the regression task.

## 9.5 *Appendix E: Sensitivity*

The case representing a typical design situation has served to perform local sensitivity analysis according to Menberg et al. [65]. Its parameters are listed in Table 6. A variation $\Delta_i$ of ±5% per design parameter has been considered to calculate local sensitivities absolute means $\mu^*$ of the Elementary Effects (EE). The sensitivities means are calculated per flow for each variable in $\boldsymbol{x}$. They have the same units as component outputs, i.e. $W_{avg}$, $W$, and $kWh/m^2a$.

**Table 6 Parameters for the representative case**

| Parameter | Unit | Mean |
|---|---|---|
| Length/Width | m | 27.5 |
| Height | m | 3.5 |
| Orientation | Degrees | 22.5 |
| Number of Floors | - | 4 |
| u-value (Wall) | | 0.2 |
| u-value (Ground Floor) | | 0.2 |
| u-value (Roof) | $W/m^2K$ | 0.2 |
| u-value (Internal Floor) | | 0.5 |
| u-value (Windows) | | 0.85 |
| g-Value (Windows) | - | 0.45 |
| Heat Capacity (Slab) | $J/m^3K$ | 900 |
| Internal Mass | $kJ/m^2K$ | 90 |
| Permeability | $m^3/m^2h$ | 7.5 |
| WWR[1] | - | 0.3 |
| Boiler Efficiency | | 0.95 |
| Heating COP | - | 3.5 |
| Cooling COP | | 3.5 |
| Operating Hours | h | 11 |
| Light Heat Gain | $W/m^2$ | 8 |
| Equipment Heat Gain | | 12 |
| Occupancy | $Person/m^2$ | 20 |

## 9.6 *Appendix F: Decision Trees*

For the local explanation modelling, we chose classification and regression trees (CART) [86] as the surrogate model and visualized the tree nodes with data distributions. As a machine learning model based upon binary trees, the decision tree naturally offers a straightforward interpretation and rule extraction structure for model interpretation. The training captures the relationship by examining and splitting data into binary hierarchical trees of interior nodes and leaf nodes. Each leaf in the decision tree is responsible for making a specific prediction. By exhaustive search, a decision tree carves up the feature space into groups of observations that share similar target values. Each leaf represents one of these groups. The order of the tree splitting is based on the "best" decision attribute for the next node, which is usually evaluated by the information gain (entropy) [87]. As implementations, the decision tree regressor from *scikit-learn* [88] and *dtreeviz* [89] for decision tree visualization have been used. To filter out the irrelevant variation impact from the value range, a min-max scaler transforms each feature individually from the original data to a range between 0 and 1 for splitting. For engineering interpretation, this scale is reverted after tree generation so that all split points and distribution diagrams include engineering units. For rule extraction in tree interpretation, special attention is paid to the selection of splitting dependent on the previous splitting. If the following split criterion differs in terms of feature selection dependent on a previous split, this is a strong indicator that an engineering rule is underlying, such as different treatment of windows size and façade insulations dependent on orientation in the example.